\documentclass[twoside]{IEEEtran}

\usepackage[T1]{fontenc}% optional T1 font encoding

\ifCLASSINFOpdf
\else
\fi
\usepackage{amsmath}
\usepackage[cmintegrals]{newtxmath}
\usepackage{enumitem}
\usepackage{booktabs}       % professional-quality tables
\usepackage{microtype}
\usepackage{array,multirow}
\usepackage{graphicx}
\usepackage[tight]{subfigure}

\usepackage{url}
\usepackage{xcolor}

\usepackage{algorithm}
\usepackage{algorithmic} 

\newcounter{ALC@tempcntr}% Temporary counter for storage

\newcommand{\NEWSECTION}[1]{%
    \setcounter{ALC@tempcntr}{\arabic{ALC@rem}}% Store old counter
    \setcounter{ALC@rem}{1}% To avoid printing line number
    \item {  \bf  #1 }  % Display comment + does not increment list item counter
    \setcounter{ALC@rem}{\arabic{ALC@tempcntr}}% Restore old counter
}%

\usepackage{amsfonts}
\usepackage{bm}

\DeclareMathOperator{\bX}{\mathbf{X}}
\DeclareMathOperator{\bXq}{\overline{\mathbf{X}}}
\DeclareMathOperator{\bxq}{\overline{\mathbf{x}}}
\DeclareMathOperator{\bWq}{\overline{\mathbf{W}}}
\DeclareMathOperator{\bw}{\mathbf{w}}
\DeclareMathOperator{\bW}{\mathbf{W}}
\DeclareMathOperator{\fF}{\mathbb{F}}

\DeclareMathOperator{\bq}{\mathbf{q}}

\DeclareMathOperator{\gC}{\nabla C}
\DeclareMathOperator{\bXtildeT}{\widetilde{\mathbf{X}}_\mathcal{T}}
\DeclareMathOperator{\bWtildeT}{\widetilde{\mathbf{W}}^{(t)}_\mathcal{T}}

\usepackage[noadjust]{cite}

\usepackage{amsthm}

\newtheorem{theorem}{Theorem}
\newtheorem{lemma}{Lemma}
\newtheorem{remark}{Remark}

\newcommand{\cpml}{CodedPrivateML}

\newcommand{\RNum}[1]{\uppercase\expandafter{\romannumeral #1\relax}}

\begin{document}
\title{{\cpml}: A Fast and Privacy-Preserving Framework for Distributed Machine Learning}

% \author{\vspace{0.2in}\onehalfspacing\IEEEauthorblockN{Jinhyun So\IEEEauthorrefmark{1} \quad  Ba\c{s}ak~G{\"u}ler\IEEEauthorrefmark{7} \quad 
% A. Salman Avestimehr\IEEEauthorrefmark{1}
% } \vspace{0.6cm}\\
% \IEEEauthorblockA{\IEEEauthorrefmark{1}University of Southern California\\ Department of Electrical and Computer Engineering\\
% Los Angeles, CA 90089 \\
% {\em jinhyuns@usc.edu, \quad avestimehr@ee.usc.edu}} \vspace{0.5cm}\\
% \IEEEauthorblockA{\IEEEauthorrefmark{7}University of California, Riverside\\
% Department of Electrical and Computer Engineering\\
% Riverside, CA 92521 \\
% {\em bguler@ece.ucr.edu}}}

\author{Jinhyun So\IEEEauthorrefmark{1},\quad Ba\c{s}ak~G{\"u}ler\IEEEauthorrefmark{1}, \quad and \quad A. Salman Avestimehr% <-this % stops a space
\thanks{
    \IEEEauthorrefmark{1} Equal contribution.
    
    This material is based upon work supported by Defense Advanced Research Projects Agency (DARPA) under Contract No. HR001117C0053, ARO award W911NF1810400, NSF grants CCF-1703575 and CCF-1763673,  and ONR Award No. N00014-16-1-2189, and a gift from Intel. The views, opinions, and/or findings expressed are those of the author(s) and should not be interpreted as representing the official views or policies of the Department of Defense or the U.S. Government. 
    
    Jinhyun So is with the Department of Electrical and Computer Engineering, University of Southern California, Los Angeles, CA, 90089 USA (e-mail: jinhyuns@usc.edu). Ba\c{s}ak G{\"u}ler is with the Department of Electrical and Computer Engineering, University of California, Riverside, CA, 92521 USA (email: bguler@ece.ucr.edu). 
    A. Salman Avestimehr is with the Department of Electrical and Computer Engineering, University of Southern California, Los Angeles, CA, 90089 USA (e-mail: avestimehr@ee.usc.edu).
    
    This work is published in IEEE Journal on Selected Areas in Information Theory \cite{so2021codedprivateml}.
    
    % This article has supplementary downloadable material available at https://doi.org/xx.xxxx/JSAIT.2021.yyyyyy, provided by the authors.
    }
}
\maketitle

\begin{abstract}
How to train a machine learning model while keeping the data private and secure? We present {\cpml}, a fast and scalable approach to this critical problem. 
{\cpml} keeps both the data and the model information-theoretically private, while allowing efficient parallelization of training across distributed workers.
We characterize {\cpml}'s privacy threshold and prove its convergence for logistic (and linear) regression. Furthermore, via extensive experiments on Amazon EC2, we demonstrate that {\cpml} provides significant speedup over cryptographic approaches based on multi-party computing (MPC).
\end{abstract}


% \fi
\IEEEpeerreviewmaketitle

% \newpage

\section{Introduction}
Modern machine learning models are breaking new ground by achieving unprecedented performance in various application domains \cite{rodrigues2019machine}. Training such models, however, is a challenging task. Due to the typically large volume of data and complexity of models, training is a compute and storage intensive task.  Furthermore,  training should often be done on sensitive data, such as healthcare records, browsing history, or financial transactions, which raises the issues of security and privacy of the dataset. This creates a challenging dilemma. On the one hand, due to its complexity, training is often desired to be outsourced to more capable computing platforms, such as the cloud. On the other hand, the training dataset is often sensitive and particular care should be taken to protect its privacy against potential breaches in such platforms. This dilemma gives rise to the main problem that we study here: {\it How can we offload the training task to a distributed computing platform, while maintaining the privacy of the dataset?}

Cloud environments often operate on a shared physical infrastructure, where multiple users share the same host machine, and are separated from each other by virtual machines that act as barriers to prevent information leakage.  
This shared environment provides significant benefits for scaling up cloud systems, but also introduces important security and privacy challenges that may result from potentially adversarial users. 
For instance, it has been shown that adversarial users can compromise the host machines by disguising themselves as regular users, and access the information of other users sharing the same host machines \cite{ristenpart2009hey, zhang2012cross, zhang2014cross, wu2014whispers, varadarajan2015placement, razavi2016flip}. 
% Moreover, attackers can even target specific users via launching co-residency attacks \cite{varadarajan2015placement}.
The focus of this paper is on privacy protection against such adversaries that can access a portion of the physical host machines in the cloud, and use them to spy on other users' datasets. 
We focus on the semi-honest adversary setup, where the adversaries follow the protocol but may leak information in an attempt to learn the training dataset. Our goal is to develop a privacy-preserving training strategy for the honest users that will protect the privacy of their datasets even if a portion of the compute machines in the cloud are controlled by adversaries.

More specifically, we consider a scenario in which a data-owner (e.g., a hospital) wishes to train a logistic regression model by offloading the large volume of data (e.g., healthcare records) and computationally-intensive training tasks (e.g., gradient computations) to $N$ machines over a cloud platform, while ensuring that any collusions between $T$ out of $N$ workers do not leak information about the training dataset. 
% We focus on the semi-honest adversary setup, where the corrupted parties follow the protocol but may leak information in an attempt to learn the training dataset. %
We propose a new framework, {\cpml} (Coded Privacy-preserving Machine Learning), towards addressing this problem. {\cpml} has three salient features:
\vspace{-0.0cm}\begin{enumerate}%[leftmargin=1cm]
\item provides strong information-theoretic privacy guarantees for both the training dataset and model parameters in the presence of colluding workers, 
\item enables fast training by distributing the  computation load effectively across several workers,
\item leverages a new method for encoding the dataset and model parameters based on coding and information theory principles, which significantly reduces the communication overhead and the complexity for distributed training. 
\vspace{-0.0cm}\end{enumerate}
% Without loss of generality, {\cpml} can also be applied to the scenario in which multiple data-owners wish to train a logistic regression model on the cloud while keeping their individual datasets private from each other and against potential collusions in the cloud. 
% In this case, no trusted party exists during training, instead, all of the training operations, including the encoding, decoding and model update operations, are performed by untrusted cloud workers. 
At a high level, {\cpml} can be described as follows. It secret shares the dataset and model parameters at each round of the training in two steps. First, it employs stochastic quantization to convert the dataset and the weight vector at each round into a finite domain. It then combines (or {\it encodes}) the quantized values with random matrices using Lagrange coding~\cite{yu2018lagrange}, to guarantee privacy (in an information-theoretic sense) while simultaneously distributing the workload among multiple workers. The challenge is however that Lagrange coding can only work for computations that are in the form of polynomial evaluations. The gradient computation for logistic regression, on the other hand, includes non-linearities that cannot be expressed as polynomials.  {\cpml} handles this challenge through polynomial approximations of the non-linear sigmoid function  in the training phase.
Upon secret sharing of the encoded dataset and model parameters, each worker performs the gradient computations using the chosen polynomial approximation, and sends the result back to the master. The workers perform the computations over the quantized and encoded data {\it as if they were computing over the uncoded dataset}. That is, the structure of the computations are the same for computing over the uncoded dataset versus computing over the encoded dataset. 
Finally, the master collects the results from a subset of fastest workers and decodes the gradient over the finite field. It then converts the decoded gradients to the real domain, updates the weight vector, and secret shares it with the worker nodes for the next round. We  note that since the computations are performed in a finite domain while the weights are updated in the real domain, the update process may lead to  undesired behaviour as weights may not converge.   Our system guarantees convergence through a stochastic quantization technique while converting between real and finite fields.

We theoretically prove that {\cpml} guarantees the convergence of the model parameters, while providing information-theoretic privacy for the training dataset. Our theoretical analysis also identifies a trade-off between privacy and parallelization. More specifically, each additional worker can be utilized either for more privacy, by  protecting against a larger number of collusions $T$, or more parallelization, by  reducing the computation load at each worker.  We characterize this trade-off for {\cpml}.
Furthermore, we empirically demonstrate the impact of {\cpml} by comparing it with the cryptographic approach based on secure multi-party computing (MPC)~\cite{yao1982protocols, ben1988completeness, beerliova2008perfectly, damgaard2007scalable}, that can also be applied to enable privacy-preserving machine learning tasks (e.g., see~\cite{nikolaenko2013privacy, gascon2017privacy, mohassel2017secureml,lindell2000privacy, dahl2018private,chen2019secure}). In particular, we envision a master who secret shares its data and model among multiple workers who collectively perform the gradient computation using a multi-round MPC protocol.  
% Given our focus on information-theoretic privacy, the most relevant MPC-based schemes for empirical comparison are the protocols from \cite{ben1988completeness, beerliova2008perfectly, damgaard2007scalable} based on Shamir's secret sharing~\cite{shamir1979share}. 
Given our focus on information-theoretic privacy, the most relevant MPC-based schemes for empirical comparison are the protocols from \cite{ben1988completeness} and  \cite{beerliova2008perfectly, damgaard2007scalable} based on Shamir's secret sharing~\cite{shamir1979share}. 
% Given our focus on information-theoretic privacy, the most relevant MPC-based scheme for empirical comparison is the recent MPC protocol from \cite{beerliova2008perfectly, damgaard2007scalable} based on Shamir's secret sharing~\cite{shamir1979share}. 
While several more recent works design MPC-based learning setups with information-theoretic privacy, their constructions are limited to three or four parties~\cite{cryptoeprint:2018:442,mohassel2018aby}.

We run extensive experiments over the Amazon EC2 cloud platform to empirically demonstrate the  performance of {\cpml}. We train a logistic regression model for image classification over the  CIFAR-10~\cite{krizhevsky2009learning} and GISETTE \cite{NIPS2004_2728} datasets, while the computation workload is distributed to up to $N=50$ machines over the cloud. We demonstrate that {\cpml} can provide significant speedup in the training time against the state-of-the-art MPC baseline (up to 5.2$\times$), while guaranteeing comparable levels of accuracy. 
This is primarily due to conventional MPC protocols' reliance on extensive communication and coordination between the workers for private computing, and not benefiting from parallelization. They can however guarantee a higher privacy threshold (i.e., larger $T$) compared with {\cpml}. 

\subsection{Other related works}
% \noindent \textbf{Other related works.} 
Apart from the MPC-based schemes, one can consider two other solutions to this problem. One is based on Homomorphic Encryption (HE)~\cite{gentry2009fully} which allows for computations to be performed on encrypted data, and has been used for privacy-preserving machine learning  solutions~\cite{gilad2016cryptonets,hesamifard2017cryptodl, graepel2012ml, yuan2014privacy, chabanne2017privacy,li2017privacy, Kim2018, 8325493, logiscticHE}. The privacy guarantees of HE are based on computational assumptions, whereas {\cpml} provides strong information-theoretic privacy. Moreover, HE  requires computations to be performed on encrypted data which leads to many orders of magnitude slow down in training. For example, for image classification on the simple MNIST dataset,  HE takes $2$ hours to  learn a logistic regression model with $96\%$  accuracy \cite{logiscticHE}, whereas for the same training setup, {\cpml} takes only $37$ seconds. This is due to the fact that, in {\cpml} there is no slow down in performing coded computations which allows for a  faster implementation.  As a trade-off, HE allows collusions between a larger number of workers  whereas in {\cpml} this number is determined by other system parameters such as the number of workers and the computation load assigned per worker.

Another possible solution is based on differential privacy (DP), which is a noisy release mechanism that preserves the privacy of personally identifiable information, in that the removal of any single element from the dataset does not change the computation outcomes significantly \cite{dwork2006calibrating}. In the context of machine learning, DP is mainly used for training when the model parameters are to be released for public use, to ensure that the individual data points from the dataset cannot be identified from the released model \cite{chaudhuri2009privacy, shokri2015privacy, abadi2016deep, pathak2010multiparty, brendan2018learning, pmlr-v22-rajkumar12, jayaraman2018distributed}. 
The main difference between these approaches and our work is that our focus is on ensuring strong information-theoretic privacy (that leaks no information about the dataset) during training, while preserving the accuracy of the  model. 
We note, however, that if the intention is to publicly release the model after training, it is in principle possible to compose the  techniques of {\cpml} with differential privacy to obtain the best of both worlds.

\section{System Model}\label{section:problem_setting}

We study the problem of training a logistic regression model. The training dataset is represented by a matrix $\mathbf{X}\in \mathbb{R}^{m\times d}$ consisting of $m$ data points with $d$ features and a label vector $\mathbf{y}\in \{0,1\}^m$. 
% Row $i$ of $\mathbf{X}$ is denoted by $\mathbf{x}_i$. 
%
The model parameters (weights) $\mathbf{w}\in \mathbb{R}^d$ are obtained by minimizing the cross entropy function,
\begin{equation}\label{cost}
C(\mathbf{w}) = \frac{1}{m}\sum_{i=1}^m \left ( -y_i \log \hat{y}_i  - (1-y_i) \log (1-\hat{y}_i)\right ),     
\end{equation} 
where $\hat{y}_i = s(\mathbf{x}_i \cdot \mathbf{w})\in (0,1)$ is the estimated probability of label $i$ being equal to $1$, $\mathbf{x}_i$ is the $i^{th}$ row of $\mathbf{X}$, 
and $s(\cdot)$ is the sigmoid function $s(z) =1/(1+e^{-z})$. 
% \begin{equation}\label{eq:sigmoid}
% g(z) =1/(1+e^{-z}). 
% \end{equation} 
The problem in \eqref{cost} can be solved via gradient descent, through an iterative process that updates the model parameters in the opposite direction of the gradient. The gradient for \eqref{cost} is given by $\nabla C (\bw) = \frac{1}{m}\mathbf{X}^{\top}(s(\mathbf{X} \times \mathbf{w}) -\mathbf{y})$. Accordingly, model parameters are updated as,
\begin{equation}\label{grad}
\mathbf{w}^{(t+1)}  =  \mathbf{w}^{(t)}-\frac{\eta}{m}\mathbf{X}^{\top}(s(\mathbf{X} \times \mathbf{w}^{(t)}) -\mathbf{y}),  
\end{equation}
where $\mathbf{w}^{(t)}$ holds the estimated parameters from iteration $t$,  $\eta$ is the learning rate, and function $s(\cdot)$ operates element-wise over the vector given by $\mathbf{X}\times \mathbf{w}^{(t)}$. 

\begin{figure}
\centering
\includegraphics[width=0.7\linewidth]{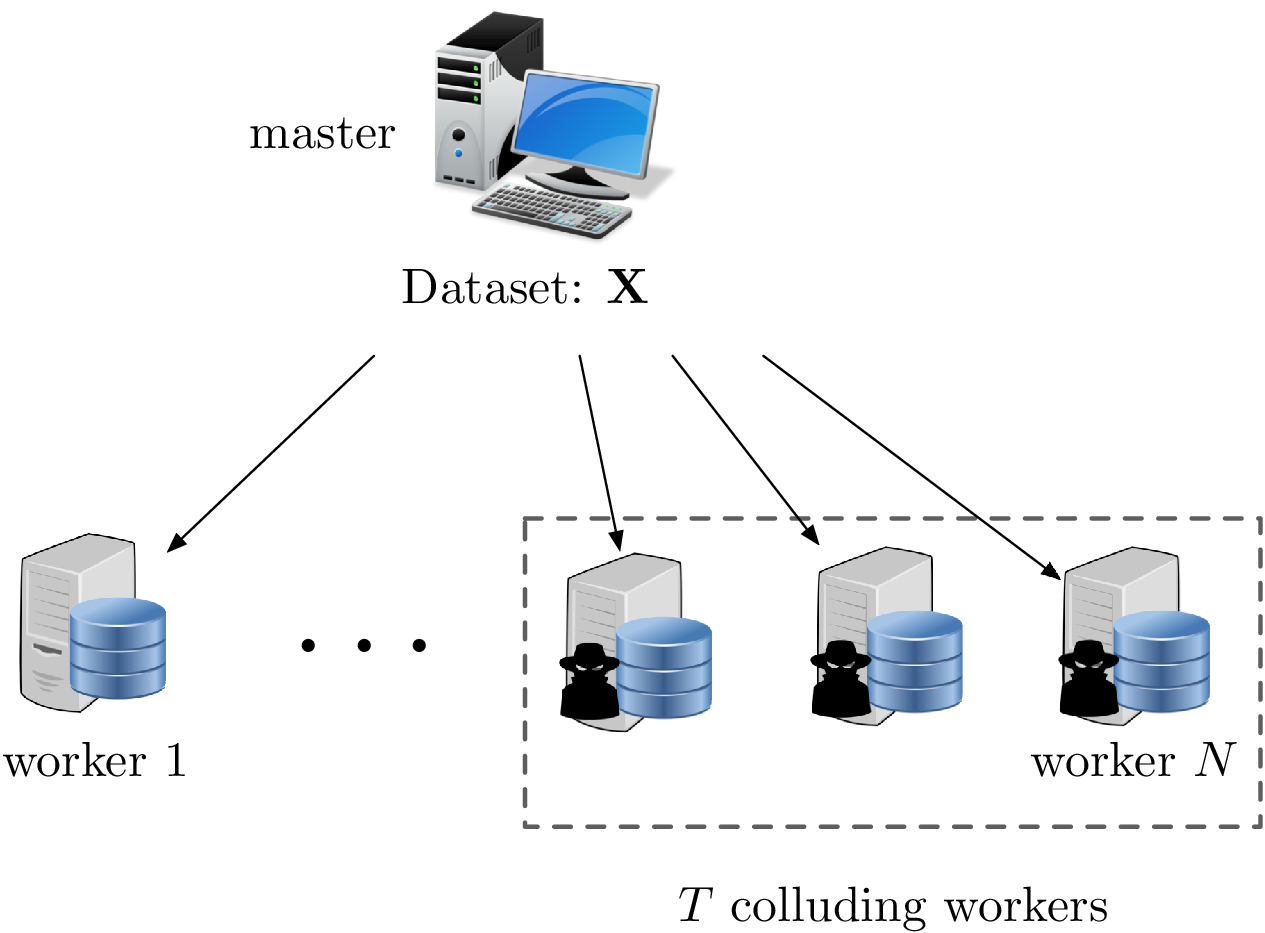}
\vspace{-0.2cm}\caption{The distributed training setup consisting of a master and $N$ worker nodes. 
}
\label{system}
\vspace{-0.5cm}\end{figure}

We consider the master-worker distributed computing architecture shown in Figure~\ref{system}, in which the master offloads the computationally-intensive operations to $N$ workers. For the training problem, these operations correspond to gradient computations in \eqref{grad}. 
% \begin{figure}
% \centering
% \includegraphics[width=0.55\linewidth]{figures/systemv7.eps}
% \vspace{-0.2cm}\caption{The distributed training setup consisting of a master and $N$ worker nodes. The master shares with each worker a coded version of dataset (denoted by $\widetilde{\mathbf{X}}_i$'s) and the current estimate of the model parameters (denoted by $\widetilde{\mathbf{W}}_{i}^{(t)}$'s) to guarantee the information-theoretic privacy of the dataset against any $T$ colluding workers. Workers perform computations locally over the coded data and send the results back to the master.  
% }
% \label{system}
% \vspace{-0.5cm}\end{figure}
In doing so, the master wishes to protect the privacy of the dataset $\mathbf{X}$ against any potential collusions between up to $T$ workers, where $T$ is  the {\it privacy parameter} of the system. 

In this work, we consider strong information-theoretic privacy, where any subset of $T$ colluding workers can not learn any information about the original dataset $\mathbf{X}$. Formally, for every subset of workers $\mathcal{T} \subseteq [N]$ of size at most $T$, we require $I\big(\bX ; \mathbf{Z}_\mathcal{T}\big) = 0$ for any distribution on $\bX$, 
where $I$ is the mutual information, and $\mathbf{Z}_\mathcal{T}$ represents the collection of all the information received by the workers in set $\mathcal{T}$ during training. The distribution of $\mathbf{X}$ may be known to the workers.  
We refer to a protocol that guarantees privacy against $T$ colluding workers as a $T$-private protocol. 
In the sequel, we present a novel protocol, {\cpml}, to solve \eqref{cost} while preserving the information-theoretic privacy of the dataset against up to $T$ colluding workers. 

\begin{remark}\normalfont
% {\cpml} can also be applied to the simpler linear regression with minor modifications.
Although our presentation is based on logistic regression, {\cpml} can also be applied to the simpler linear regression model with minor modifications.
\end{remark}

\section{The {\cpml} Protocol}\label{section:proposed_scheme}

% \begin{figure}
% \centering
% \includegraphics[width=0.55\linewidth]{figures/CPML_flowchart.eps}
% \vspace{-0.2cm}\caption{Flowchart of {\cpml}.  
% }
% \label{fig:flowchart}
% \vspace{-0.5cm}\end{figure}

% {\cpml} consists of the following four main phases.  

% {\cpml} consists of four main components that are first described at a high-level below, and then with details in the rest of this section.  

\begin{figure}
\centering
\includegraphics[width=0.7\linewidth]{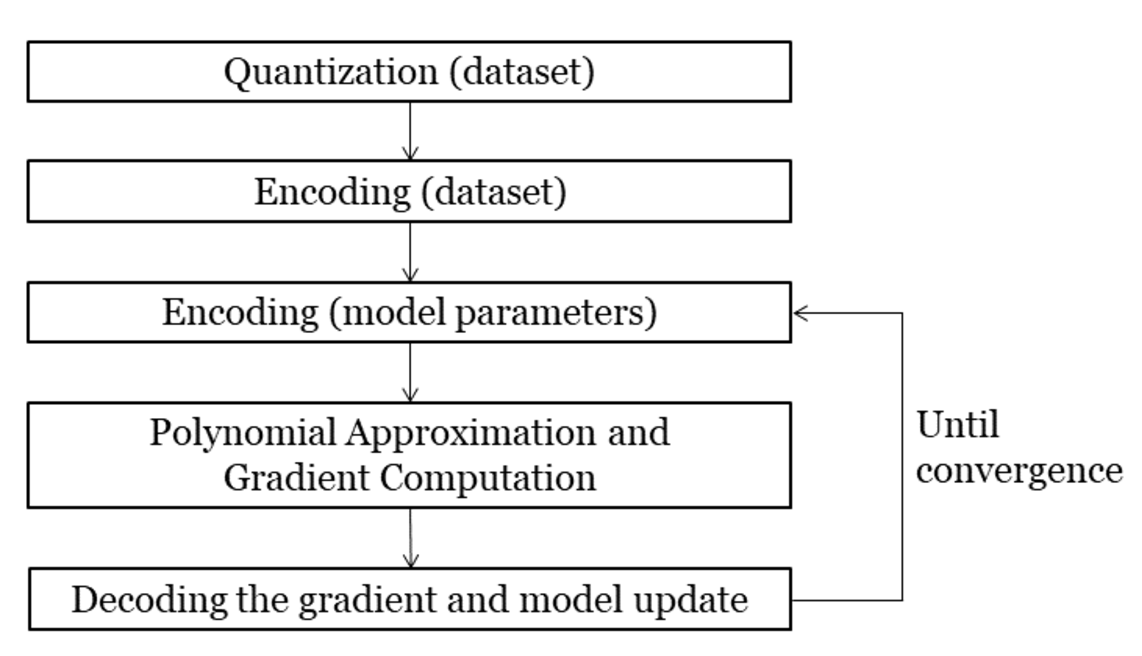}
\vspace{-0.2cm}\caption{Flowchart of {\cpml}.  
}
\label{fig:flowchart}
\vspace{-0.5cm}\end{figure}

% In this section, we present the {\cpml} protocol.
{\cpml} consists of four main components: 1) quantization, 2) encoding, 3) polynomial approximation and gradient computation, and 4) decoding the gradient and model update. Figure~\ref{fig:flowchart} shows the flowchart of {\cpml}. In the first component, the master quantizes the dataset from the real domain to the domain of integers, and then embeds it in a finite field. 
In the second component, the master encodes the quantized dataset and sends them to the workers.
At each iteration, the master also quantizes and encodes the model parameters. 
In the third component, given the encoded dataset and model parameters, each worker performs the gradient computations by using polynomial approximation to substitute the sigmoid function. 
In the last component, the master decodes the gradient computations and converts them from the finite field to real domain, and updates the model parameters in the real domain. 
This process is iterated until the model parameters converge. 

We now provide the details of each component.

\vspace{-0.3cm}\subsection{Quantization}\label{Sec:quan}
In order to guarantee information-theoretic privacy, one has to mask  the dataset and weights in a finite field\footnote{We need a finite field instead of a ring as our encoding and decoding schemes based on Lagrange coding, which we explain in Sections \ref{Sec:encoding} and \ref{sec:decoding}, require division (or inverse multiplication) which the ring does not have in general.} $\fF$ using uniformly random matrices, so that the added randomness can make each data point appear equally likely. In contrast, the dataset and weights for the training task are defined in the domain of real numbers. Our solution to handle the conversion between the real and finite domains is through the use of stochastic quantization. Accordingly, in the first component of our system, master quantizes the dataset and weights from the real domain to the domain of integers, and then embeds them in a field $\mathbb{F}_p$ of integers modulo a prime $p$. 
The quantized version of the dataset $\mathbf{X}$ is given by $\overline{\mathbf{X}}$. 
The quantization of the weight vector $\mathbf{w}^{(t)}$, on the other hand, is represented by a matrix $\overline{\mathbf{W}}^{(t)}$, where  each  column holds an independent stochastic quantization of $\mathbf{w}^{(t)}$. 
This structure will be important for the convergence of the model.

We consider an element-wise lossy quantization scheme for the dataset and weights. For quantizing the dataset $\mathbf{X} \in \mathbb{R}^{m \times d}$, we use a simple deterministic rounding technique: 
\begin{equation}\label{eq:round} 
    Round(x) =
    \left\{
    \begin{array}{ll}
          \lfloor x \rfloor & \text{if \quad } x-\lfloor x \rfloor < 0.5 \\
          \lfloor x \rfloor+1 & \text{otherwise } 
    \end{array} 
    \right. ,
\end{equation}
where $\lfloor x \rfloor$ is the largest integer less than or equal to $x$.
We define the quantized dataset as
\begin{equation}\label{eq:X_q} 
\overline{\bX}\triangleq \phi\left(Round(2^{l_x} \cdot \mathbf{X})\right),
\end{equation}
where the rounding function from \eqref{eq:round} is applied element-wise to the elements of matrix $\bX$ and $l_x$ is an integer parameter that controls the quantization loss.  Function $\phi:\mathbb{R}\rightarrow\mathbb{F}_{p}$ is a mapping defined to represent a negative integer in the finite field by using two's complement representation,
\begin{equation}\label{eq:phi} 
    \phi(x) =
    \left\{
    \begin{array}{ll}
          x & \text{if } x \geq 0\\
          p+x & \text{if } x<0
    \end{array} 
    \right. .
\end{equation}
%\textbf{Salman: There is a typo here -- both end points are the same. please fix carefully} 
Note that the domain of \eqref{eq:X_q} is $\big[-\frac{p-1}{2^{(l_x+1)}},\frac{p-1}{2^{(l_x+1)}}\big]$. To avoid a wrap-around which may lead to an overflow error, prime $p$ should be large enough, i.e., $p\geq 2^{l_x+1}\max \{ \lvert{\bX}_{i,j}\rvert \} +1$.  
The value of $p$ also depends on the bitwidth of the machine as well as the number of features $d$.
For instance, in our experiments presented in  Section~\ref{section:experiments}, we select $p=2^{25}-37$ in a $64$-bit implementation with the GISETTE dataset whose number of features is $d=5000$. This is the largest prime to avoid an overflow on intermediate multiplications. More specifically, we do a modular operation after the inner product of vectors instead of doing a modular operation per product of each element in order to speed up the running time of matrix-matrix multiplication.
To avoid an overflow on this, $p$ should satisfy $d(p-1)^2 \leq 2^{64} -1$.
% To avoid an overflow on this, $p$ should be smaller than a threshold given by $d(p-1)^2 \leq 2^{64} -1$.

At each iteration $t$, master also quantizes the weight vector $\mathbf{w}^{(t)}$ from real domain to the finite field.
% At each iteration $t$, master also quantizes the weight vector $\mathbf{w}^{(t)}$ from real domain to the finite field\footnote{\color{blue}The weight vector $\bw^{(0)}$ is initialized as a Gaussian random vector with zero mean and unit variance.}. 
This proves to be a challenging task as it should be performed in a way to  ensure the convergence of the model. Our solution to this is a quantization technique inspired by \cite{pmlr-v70-zhang17e, ZipML_arXiv}. 
Initially, we define a stochastic quantization function:
\begin{equation}\label{eq:quan} 
Q(x;l_w) \triangleq \phi\big(Round_{stoc}(2^{l_w} \cdot x)\big),
\end{equation}
where $l_w$ is an integer parameter to control the quantization loss.  
$Round_{stoc}:\mathbb{R}\rightarrow\mathbb{R}$ is a stochastic rounding function: 
\begin{equation*}
    Round_{stoc}(x) =
    \left\{
    \begin{array}{ll}
          \lfloor x \rfloor & \text{with prob. } 1-(x-\lfloor x \rfloor)\\
          \lfloor x \rfloor+1 & \text{with prob. } x-\lfloor x \rfloor 
    \end{array} 
    \right. . 
\end{equation*}
The probability of rounding $x$ to $\lfloor x \rfloor$ is proportional to the proximity of $x$ to $\lfloor x \rfloor$ so that stochastic rounding is unbiased (i.e., $\mathbb{E}[Round_{stoc}(x)]=x$). 

For quantizing the weight vector $\mathbf{w}^{(t)}$,  the master creates $r$ independent quantized vectors: 
\begin{equation}\label{eq:vector}
\overline{\mathbf{w}}^{(t), j}\triangleq Q_j(\mathbf{w}^{(t)};l_w)\in \mathbb{F}^{d\times1}_p \text{\;\; for \;\;} j\in[r],
\end{equation}
where the quantization function \eqref{eq:quan} is applied element-wise to the vector $\mathbf{w}^{(t)}$ and each $Q_j(\cdot; \cdot)$ denotes an independent realization of \eqref{eq:quan}. 
To avoid a wrap-around which may lead to an overflow error, prime $p$ should be large enough, i.e., $p\geq 2^{l_x+1}\max \{ \lvert{\bw}^{(t)}_{i}\rvert \} +1$.
% Similar to the quantized dataset, to avoid a wrap-around which may lead to an overflow error, prime $p$ should be large enough, i.e., $p\geq 2^{l_x+1}\max \{ \lvert{\bw}^{(t)}_{i}\rvert \} +1$.
The number of quantized vectors $r$ is equal to the degree of the polynomial approximation for the sigmoid function, which we will describe later in Section~\ref{sec:computing}. The intuition behind creating $r$ independent quantizations is to ensure that the gradient computations performed using the quantized weights are unbiased estimators of the true gradients. As detailed in Section~\ref{section:performance_analysis}, this property is fundamental for the convergence analysis of our model. 
% We also note that as the number $r$ depends only on the degree of the polynomial approximation and is a fixed value that does not increase with respect to other system parameters.  
The specific values of parameters $l_x$ and $l_w$ provide a trade-off between the rounding error and overflow error. In particular, a larger value reduces the rounding error while increasing the chance of an overflow. We denote the quantization of the weight vector $\mathbf{w}^{(t)}$ as
\begin{equation}\label{eq:quantized} 
\overline{\mathbf{W}}^{(t)} = [\overline{\mathbf{w}}^{(t), 1} \;\; \cdots \;\; \overline{\mathbf{w}}^{(t), r}], 
\end{equation}
by arranging the quantized vectors from \eqref{eq:vector} in matrix form.

\vspace{-0.3cm}
\subsection{Encoding the Dataset and the Model}\label{Sec:encoding}
% \subsection{Encoding and Secret Sharing}\label{Sec:encoding}
In the second component, the master partitions the quantized dataset $\overline{\mathbf{X}} \in \mathbb{F}^{m \times d}_{p}$ into $K$ submatrices and encodes them using Lagrange coding~\cite{yu2018lagrange}. 
It then sends to worker $i\in[N]$ a coded submatrix $\widetilde{\mathbf{X}}_{i} \in \mathbb{F}^{\frac{m}{K} \times d}_{p}$.
This encoding enables two salient features of {\cpml}, parallelization and information-theoretic privacy guarantees.
First, parameter $K$ is related to the computation load at each worker (i.e., what fraction of the dataset is processed at each worker) because the size of encoded dataset is $1/K$-th of the size of original dataset $\bX$. 
As we will show later, we can increase the parameter $K$ as $N$ increases, which reduces the computation overhead of each worker and communication overhead between the master and workers. 
This property enables our approach to scale to a significantly larger number of workers than state-of-the-art privacy preserving machine learning approaches.
% This property enables our approach to scale to a significantly larger number of workers than state-of-the-art privacy preserving machine learning approaches (i.e., beyond 3-4 workers).
Second, this encoding ensures that the coded matrices do not leak any information about the original dataset $\bX$ even if $T$ workers collude, which will be showed in Section \ref{section:performance_analysis}.
In addition, the master has to ensure the weight estimations sent to the workers at each iteration do not leak information about the dataset. This is because the weights updated via \eqref{grad} carry information about the whole training set, and sending them directly to the workers may breach privacy. In order to prevent this, at iteration $t$, the master also quantizes the current weight vector $\mathbf{w}^{(t)}$ to the finite field and encodes it again using Lagrange coding.

We now state the details of our second component. The master first partitions the quantized dataset $\overline{\bX}$ into $K$ submatrices $\overline{\bX}=[\overline{\bX}^{\top}_1\ldots\overline{\bX}^{\top}_K]^{\top}$, where $\overline{\bX}_i \in \mathbb{F}^{\frac{m}{K}\times d}_{p}$ for $i\in[K]$. We assume that $m$ is divisible by $K$.
Next, the master selects $K+T$ distinct elements $\beta_1,\ldots,\beta_{K+T}$ from $\mathbb{F}_p$ and employs Lagrange coding~\cite{yu2018lagrange} to encode the dataset. To do so, the master forms a polynomial $u:\mathbb{F}_p\rightarrow \mathbb{F}_p^{\frac{m}{K}\times d}$ of degree %\blue
{at most} $K+T-1$ such that $u(\beta_i)=\overline{\mathbf{X}}_i$ for $i\in[K]$, and~$u(\beta_i)=\mathbf{R}_i$ for~$i\in\{K+1,\ldots,K+T\}$, where $\mathbf{R}_i$'s are chosen uniformly at random from~$\mathbb{F}^{\frac{m}{K}\times d}_p$ (the role of $\mathbf{R}_i$'s is to mask the dataset and provide privacy against up to $T$ colluding workers). This can be  accomplished by letting~$u$ be the respective \textit{Lagrange interpolation polynomial,}
\begin{equation}\label{eq:lag1} 
    u(z) 
    \triangleq\!\! \sum_{j\in[K]}\!\overline{\mathbf{X}}_j\cdot \!\!\!\prod_{k\in [K+T]\setminus\{j\}}\!\frac{z\!-\!\beta_k}{\beta_j\!-\!\beta_k}  \!+\! \sum_{j=K+1}^{K+T} \!\mathbf{R}_j\cdot \!\prod_{k\in [K+T]\setminus\{j\}}\!\frac{z\!-\!\beta_k}{\beta_j\!-\!\beta_k}.
\end{equation}
The master then selects $N$ distinct elements~$\{\alpha_i\}_{i\in[N]}$ from $\mathbb{F}_p$ such that $\{\alpha_i\}_{i\in[N]}\cap\{\beta_j\}_{j\in[K]}=\varnothing$, and encodes the dataset by letting $\widetilde{\bX}_i=u(\alpha_i)$ for $i\in[N]$. By defining an encoding matrix $\mathbf{U}=[\mathbf{u}_1\ldots \mathbf{u}_N]\in\mathbb{F}_p^{(K+T)\times N}$ whose $(i,j)^{th}$ element is given by 
$u_{ij}=\prod_{\ell\in[K+T]\setminus \{i\}}\frac{\alpha_j-\beta_\ell}{\beta_i-\beta_\ell}$, one can also represent the encoding of the dataset as
\begin{equation}\label{equation:perWorkerEncoding} 
     \widetilde{\mathbf{X}}_i=u(\alpha_i)=(\overline{\mathbf{X}}_1,\ldots,\overline{\mathbf{X}}_K,\mathbf{R}_{K+1},\ldots,\mathbf{R}_{K+T})\cdot \mathbf{u}_i.
\end{equation} 
At iteration $t$, the quantized weights  $\overline{\mathbf{W}}^{(t)}$ are also encoded using a Lagrange interpolation polynomial,
\begin{equation}\label{eq:lag2}
    v(z) 
    \triangleq\!\! \sum_{j\in[K]}\!\overline{\mathbf{W}}^{(t)}\cdot \!\!\!\prod_{k\in [K+T]\setminus\{j\}}\!\frac{z\!-\!\beta_k}{\beta_j\!-\!\beta_k} +\!\sum_{j=K+1}^{K+T} \!\mathbf{V}_j\cdot \!\prod_{k\in [K+T]\setminus\{j\}}\!\frac{z\!-\!\beta_k}{\beta_j\!-\!\beta_k},
\end{equation}
%\begin{align}\label{equation:perWorkerEncoding_w}
%     \widetilde{\mathbf{w}}_i\!=\!v(\alpha_i)\!=\!(\overline{\mathbf{w}}^{(t)},\ldots,\overline{\mat%hbf{w}},\mathbf{v}_{K+1},\ldots,\mathbf{v}_{K+T})\cdot U_i,
%\end{align} 
where $\mathbf{V}_j$ for $j\in[K+1,K+T]$ are chosen uniformly at random from~$\mathbb{F}^{d\times r}_p$. The coefficients $\beta_1, \ldots, \beta_{K+T}$ are the same as in  \eqref{eq:lag1}, and we have the property  $v(\beta_{i})=\overline{\mathbf{W}}^{(t)}$ for  $i\in [K]$. 
The master then encodes the quantized weight vector by using  the same evaluation points~$\{\alpha_i\}_{i\in[N]}$. Accordingly, the weight vector is encoded as 
\begin{equation}\label{equation:weightencoding} 
    \widetilde{\mathbf{W}}^{(t)}_i\!=\!v(\alpha_i)\!=\!(\overline{\mathbf{W}}^{(t)}\!\!,\ldots,\overline{\mathbf{W}}^{(t)}\!\!,\mathbf{V}_{K+1},\ldots,\mathbf{V}_{K+T})\cdot \mathbf{u}_i,
\end{equation}  
for $i\in [N]$, using the encoding matrix $\mathbf{U}$ from  \eqref{equation:perWorkerEncoding}. 
The degree of the polynomials $u(z)$ and $v(z)$ are both $K+T-1$.

\subsection{Polynomial Approximation and Gradient Computation}\label{sec:computing}
% {\color{blue} In the third component, each worker locally performs the gradient computations and sends the result back to the master. We note that the workers perform the computations over the encoded data as if they were computing over the true dataset. That is, the structure of the computations are the same for computing over the true dataset versus computing over the encoded dataset.  
% A major challenge is that Lagrange coding is designed for distributed polynomial computations. However, the computations in the training phase are not polynomials due to the sigmoid function.  We overcome this by approximating the sigmoid with a polynomial of a selected degree $r$. This allows us to represent the gradient computations in terms of polynomials that can be computed locally by each worker. 
% {\cpml} guarantees the convergence to the optimal loss function $C(\bw^*)$ where $C$ is the cross entropy function defined in \eqref{cost}, even though we use the polynomial approximation to substitute the sigmoid function in the update equation \eqref{grad}, which will be demonstrated in our theoretical results in Section \ref{section:performance_analysis}.}

Upon receiving the encoded (and quantized) dataset and weights, workers should proceed with gradient computations. However, a major challenge is that Lagrange coding is originally designed for polynomial computations, while the gradient computations are not polynomials due to the sigmoid function. Our solution is to use a polynomial approximation of the sigmoid function,
\begin{equation}\label{eq:poly_approximation} 
    \hat{s}(z) = \sum_{i=0}^{r} c_{i} z^{i},
\end{equation}
where $r$ and $c_i$ denotes the degree and coefficients of the polynomial, respectively. The coefficients are obtained by fitting the sigmoid function via least squares estimation. Using this polynomial approximation we can rewrite \eqref{grad} as
\begin{equation}\label{eq:apprx_grad1} 
    \bw^{(t+1)}  =   \bw^{(t)}-\frac{\eta}{m}\overline{\mathbf{X}}^{\top}(\hat{s}(\overline{\mathbf{X}} \times \mathbf{w}^{(t)}) -\mathbf{y}).
\end{equation} 
where $\overline{\mathbf{X}}$ is the quantized version of $\mathbf{X}$, and  $\hat{s}(\cdot)$ operates element-wise over the vector $\overline{\mathbf{X}} \times \mathbf{w}^{(t)}$. 

Another challenge is to ensure the  convergence of weights. As we detail in Section~\ref{section:performance_analysis}, this necessitates the gradient estimations to be unbiased  using the polynomial approximation with quantized weights. We solve this by utilizing the  computation technique from Section~$4.1$ in \cite{ZipML_arXiv} using the quantized weights formed in Section~\ref{Sec:quan}. 
Specifically, given a degree $r$ polynomial from \eqref{eq:poly_approximation} and $r$ independent quantizations from \eqref{eq:quantized}, we define a function
\begin{equation}\label{eq:app} 
\bar{s}(\overline{\mathbf{X}}, \overline{\mathbf{W}}^{(t)})  \triangleq \sum_{i=0}^r c_i \prod_{j\leq i} (\overline{\mathbf{X}}\times \overline{\mathbf{w}}^{(t), j}),
\end{equation}
where the product $\prod_{j\leq i}$ operates element-wise over the vectors $(\overline{\mathbf{X}}\times \overline{\mathbf{w}}^{(t), j})$ for $j\leq i$.  
Lastly, we note that \eqref{eq:app} is an unbiased estimator of  $\hat{s}(\overline{\mathbf{X}}\times \mathbf{w}^{(t)})$,
\begin{equation}\label{eq:unbias} 
E[\bar{s}(\overline{\mathbf{X}}, \overline{\mathbf{W}}^{(t)}) ] =\hat{s}(\overline{\mathbf{X}}\times \mathbf{w}^{(t)}), 
\end{equation}
where $\hat{s}(\cdot)$ acts element-wise over the vector $\overline{\mathbf{X}} \times \mathbf{w}^{(t)}$, and the result follows from the independence of quantizations. 
% We can now represent the computations from \eqref{grad} in terms of a polynomial. That is,
Using \eqref{eq:app}, we rewrite the update equations from \eqref{eq:apprx_grad1} using  quantized weights,
\begin{equation}\label{eq:apprx_grad} 
\bw^{(t+1)}  =   \bw^{(t)}-\frac{\eta}{m}\overline{\mathbf{X}}^{\top}(\bar{s}(\overline{\mathbf{X}}, \overline{\mathbf{W}}^{(t)}) -\mathbf{y}).
\end{equation} 
{\cpml} guarantees the convergence to the optimal loss function $C(\bw^*)$ where $C$ is the cross entropy function defined in \eqref{cost}, even though we use the polynomial approximation to substitute the sigmoid function in the update equation \eqref{grad}, which will be demonstrated in Section \ref{section:performance_analysis}.

%Upon inspecting \eqref{eq:apprx_grad}, one can observe that the term $\overline{\mathbf{X}}^{\top}\bar{s}(\overline{\mathbf{X}}, \overline{\mathbf{W}}^{(t)})$ is a polynomial over the inputs $\overline{\mathbf{X}}$ and $\overline{\mathbf{W}}^{(t)}$, which can then be evaluated using LCC in a distributed and privacy-preserving manner. Moreover, the computations in \eqref{eq:apprx_grad} provide an unbiased estimator of \eqref{eq:apprx_grad1}, which is essential in our convergence analysis. 
Computations are then performed at each worker locally. 
At each iteration, worker~$i\in [N]$ locally computes $f:\fF^{\frac{m}{K}\times d}_{p}\times\fF^{d\times r}_{p}\rightarrow\fF^{d}_{p}$,
\begin{equation}\label{eq:compute} 
f\big(\widetilde{\bX}_i,\widetilde{\bW}^{(t)}_i\big)=\widetilde{\bX}^{\top}_i \, \bar{s}(\widetilde{\bX}_i ,  \widetilde{\bW}^{(t)}_i),
\end{equation}
using $\widetilde{\mathbf{X}}_i$ and $\widetilde{\mathbf{W}}_i^{(t)}$ and sends the result back to the master.
% As detailed in the following section, after receiving the results \eqref{eq:compute} from a sufficient number of workers, the master can decode $\Big\{ f\big(\overline{\bX}_k,\overline{\bw}^{(t)}\big)\Big\}_{k\in [K]}$ and  compute $\mathbf{X}^{\top}\hat{s}(\mathbf{X} \times \mathbf{w}^{(t)}) = \sum_{k=1}^K f\big(\overline{\bX}_k,\overline{\bw}^{(t)}\big)=\sum_{k=1}^K \overline{\bX}^{\top}_k \, \hat{s}\big(\overline{\bX}_k \times  \overline{\bw}^{(t)}\big)$. 
% Note that this computation is performed using finite field arithmetic and $\text{deg}(f)$ is $2r+1$.
This computation is a polynomial function evaluation in finite field arithmetic and the degree of $f$ is $\text{deg}(f) = 2r+1$.

% \textbf{Payman: it wasn't mentioned anywhere how the approximating polynomials was chosen? shouldn't we add some discussion of that?}

% Note that this computation is performed using finite field arithmetic and the degree of $f$ is $\text{deg}(f) = 2r+1$.

% As we detail in the following, after receiving the results \eqref{eq:compute} from a sufficient number of workers, the master can decode $\Big( f\big(\overline{\bX}_k,\overline{\bw}^{(t)}\big)\Big)_{k=1}^K$ and  compute $\mathbf{X}^{\top}\hat{s}(\mathbf{X} \times \mathbf{w}^{(t)}) = \sum_{k=1}^K f\big(\overline{\bX}_k,\overline{\bw}^{(t)}\big)=\sum_{k=1}^K \overline{\bX}^{\top}_k \, \hat{s}\big(\overline{\bX}_k \times  \overline{\bw}^{(t)}\big)$. Note that this computation is performed using finite field arithmetic and $\text{deg}(f)$ is $2r+1$.

\subsection{Decoding the Gradient and Model Update}\label{sec:decoding}
% {\color{blue}In the last component, the master collects the local computation results from a subset of  workers and decodes the gradient. Then, the master converts the gradient from finite to real domain and updates the weight vector for the next round in the real domain.}

After receiving the evaluation results in \eqref{eq:compute} from a
sufficient number of workers, master decodes 
$\Big\{ f\big(\overline{\bX}_k,\overline{\bW}^{(t)}\big)\Big\}_{k\in [K]}$ over the finite field. 
The minimum number of workers needed for the decoding operation to be successful, which we call the \emph{recovery threshold} of the protocol, is equal to $(2r+1)(K+T-1)+1$ as we demonstrate in Section~\ref{section:performance_analysis}.
% workers whose number equals to \textit{recovery threshold}$=\text{deg}(f)\cdot(K+T-1)+1$, master decodes 
% $\Big\{ f\big(\overline{\bX}_k,\overline{\bw}^{(t)}\big)\Big\}_{k\in [K]}$ over the finite field. 
% For any multivariate polynomial $f\big(\overline{\mathbf{X}}_i,\overline{\mathbf{w}}^{(t)}\big)$, 

We now proceed to the details of decoding. 
By construction of the Lagrange polynomials in \eqref{eq:lag1} and \eqref{eq:lag2}, one can  define a univariate polynomial $h(z)=f\big(u(z),v(z)\big)$ such that
\begin{equation}
 h(\beta_i)=f\big(u(\beta_i),v(\beta_i)\big)
=f\big(\overline{\bX}_i,\overline{\bW}^{(t)}\big) 
= \overline{\mathbf{X}}_i^{\top}\bar{s}(\overline{\mathbf{X}}_i, \overline{\mathbf{W}}^{(t)}), \label{eq:beta}
\end{equation}
for $i\in[K]$. On the other hand, from \eqref{eq:compute}, the computation result from worker $i$ equals to 
\begin{equation}
h(\alpha_i)=f\big(u(\alpha_i),v(\alpha_i)\big) 
=  f\big(\widetilde{\bX}_i,\widetilde{\bW}^{(t)}_i\big) 
=\widetilde{\bX}^{\top}_i \, \bar{s}(\widetilde{\bX}_i ,  \widetilde{\bW}^{(t)}_i). \label{eq:alpha}
\end{equation}
%Since $\text{deg}(h(z))\leq\text{deg}(f)\cdot(K+T-1)$, master can obtain all coefficients of $h(z)$ by obtaining $\text{deg}(f)\cdot(K+T-1)+1$ evaluation results from any worker. Having this polynomial, the master can obtain desired evaluations $f\big(\overline{\bX}_i,\overline{\bw}^{(t)}_i\big)=h(\beta_i)$ for every $i\in[K]$.  
The main intuition behind the decoding process is to use the computations from \eqref{eq:alpha} as evaluation points $h(\alpha_i)$ to interpolate the polynomial $h(z)$. 
% After $h(z)$ is recovered, the desired computations in \eqref{eq:beta} can be recovered by evaluating $h(\beta_i)$ for $i\in [K]$.  
Specifically, the master can obtain all coefficients of $h(z)$ from $(2r+1)(K+T-1)+1$ evaluation results as the degree of the polynomial $h(z)$ is less than or equal to $(2r+1)(K+T-1)$. 
After $h(z)$ is recovered, the master can recover \eqref{eq:beta} by computing $h(\beta_i)$ for  $i\in[K]$.
To do so, the master performs polynomial interpolation in a finite field. Upon receiving the local computation $f\big(\widetilde{\bX}_i,\widetilde{\bW}^{(t)}_i\big)$ in \eqref{eq:alpha} from at least $(2r+1)(K+T-1)+1$ workers, the master computes
\begin{equation}\label{eq:decoding}
    f\big(\overline{\bX}_k,\overline{\bW}^{(t)}\big) \triangleq \sum_{i\in\mathcal{I}}f(\widetilde{\bX}_i, \widetilde{\mathbf{W}}^{(t)}_i)\cdot \!\!\!\! \prod_{j\in \mathcal{I}\setminus\{i\}}\frac{\beta_k-\alpha_j}{\alpha_i-\alpha_j}
\end{equation}
for $k\in[K]$, where $\mathcal{I}\subseteq[N]$ denotes the set of the first $(2r+1)(K+T-1)+1$ workers who send their local computations  $f(\widetilde{\bX}_i, \widetilde{\mathbf{W}}^{(t)}_i)$ to the master. 
The master then aggregates the decoded computations $f\big(\overline{\bX}_k,\overline{\bW}^{(t)}\big)$ to compute the desired gradient as,
\begin{equation}
     \sum_{k=1}^K f(\overline{\bX}_k,\overline{\bW}^{(t)})
    =\sum_{k=1}^K \overline{\bX}^{\top}_k \bar{s}(\overline{\bX}_k ,  \overline{\bW}^{(t)}) 
    = \overline{\mathbf{X}}^{\top}\bar{s}(\overline{\mathbf{X}} ,  \overline{\mathbf{W}}^{(t)}). \label{eq:sum}
\end{equation}
% Lastly, master converts \eqref{eq:sum} from the finite field to real domain to obtain $\bX^{\top}\hat{s}(\bX \time \mathbf{w})$, and update the weight vector.
Lastly, master converts \eqref{eq:sum} from the finite field to the real domain and updates the weights according to \eqref{eq:apprx_grad} in the real domain.
This conversion is attained by the function
\begin{equation}\label{eq:inverse_Q}
Q^{-1}_{p}(\overline{x};l)=2^{-l}\cdot {\phi}^{-1}\big(\overline{x}\big),
\end{equation}
where we let $l = l_x + r(l_x + l_w)$, and ${\phi}^{-1}:\mathbb{F}_p\rightarrow\mathbb{R}$ is defined as,
\begin{equation}\label{eq:inv_phi}
{\phi}^{-1}(\overline{x})=
    \left\{
    \begin{array}{ll}
          \overline{x} & \text{if \quad } 0 \leq \overline{x} < \frac{p-1}{2}\\
          \overline{x}-p & \text{if \quad } \frac{p-1}{2} \leq \overline{x} < p
    \end{array} .
    \right.
\end{equation}
The overall procedure of {\cpml}  is given in Algorithm~\ref{algorithm:CodedPrivateML}.

% \subsection{Algorithms}\label{app:algorithms}
% % \input{algorithms.tex} 

% The overall procedure of the {\cpml} protocol is given in Algorithm~\ref{algorithm:CodedPrivateML}. 
% Procedures for the individual phases from  Section~\ref{section:proposed_scheme} are provided in Algorithms~\ref{alg:quan}-\ref{alg:dec}, respectively.

\begin{algorithm}[t]
  \caption{{\cpml}}\label{algorithm:CodedPrivateML} \small
  \begin{algorithmic}[1]
    \INPUT{Dataset $\bX,\mathbf{y}$} \
    \OUTPUT{Model parameters $\bw^{(J)}$ where $J$ is the number of iterations} \ 
    \vspace{0.1cm}
    \NEWSECTION{Master:}
    % \vspace{0.1cm}
    \STATE Compute the quantized dataset $\overline{\bX}$ using  \eqref{eq:X_q}. 
    \STATE Form the encoded matrices $\{\widetilde{\bX}_i\}_{i\in[N]}$ in \eqref{equation:perWorkerEncoding}. 
    \STATE Send $\widetilde{\bX}_i$ to worker $i\in[N]$.
    \STATE Initialize the weights $\bw^{(0)}\in \mathbb{R}^{d \times1}$.
    \FOR{iteration $t=0,\ldots,J-1$}
    % \NEWSECTION{Master:}
    \STATE Find the quantized weights  $\overline{\bW}^{(t)}$ from \eqref{eq:quantized}. 
    \STATE Encode $\overline{\bW}^{(t)}$ into  $\{\widetilde{\bW}_i^{(t)}\}_{i\in[N]}$ using \eqref{equation:weightencoding}. 
    \STATE Send $\widetilde{\bW}_i^{(t)}$ to worker $i\in[N]$.
    \vspace{0.1cm}
    \NEWSECTION{Workers:}
    % \vspace{0.1cm}
     \FOR{worker $i=1,\ldots,N$}
    \STATE Compute $f(\widetilde{\bX}_i,\widetilde{\bW}_i^{(t)})$ from  \eqref{eq:compute} and send the result back to the master.  
    \ENDFOR
    \vspace{0.1cm}
    \NEWSECTION{Master:}
    % \vspace{0.1cm}
    \IF{Results received from  at least $(2r+1)(K+T-1)+1$ workers}
        \STATE Decode
        $\{f(\overline{\bX}_k,\overline{\bW}^{(t)})\}_{k\in[K]}$ via polynomial interpolation from the received results.  
    \ENDIF
    \STATE Compute $\sum_{k=1}^K f(\overline{\bX}_k,\overline{\bW}^{(t)})$ in \eqref{eq:sum} and convert it from finite field to real domain using \eqref{eq:inverse_Q}. 
    \STATE Update the weights  via  \eqref{eq:apprx_grad} to obtain $\bw^{(t+1)}$. 
    \ENDFOR
    \STATE \textbf{return} $\bw^{(J)}$
  \end{algorithmic}
\end{algorithm}
\vspace{-0.4cm}

\section{Theoretical Results}\label{section:performance_analysis}

Consider the cost function \eqref{cost} when the dataset $\mathbf{X}$ is replaced with the quantized dataset $\overline{\bX}$. 
Also, denote $\bw^{*}$ as the optimal weight vector that minimizes~\eqref{cost} when $\hat{y}_i = s(\bxq_{i}\cdot \bw)$, where $\overline{\mathbf{x}}_i$ is row $i$ of $\bXq$. 
In this section, we prove that {\cpml} guarantees convergence to the optimal model parameters (i.e., $\bw^{*}$) while maintaining the privacy of the dataset against colluding workers. 
Recall that the model update at the master follows from \eqref{eq:apprx_grad}, which is
\begin{equation}\label{eq:updateEQN}
\bw^{(t+1)}  =   \bw^{(t)}-\frac{\eta}{m}\overline{\mathbf{X}}^{\top}(\bar{s}(\overline{\mathbf{X}}, \overline{\mathbf{W}}^{(t)}) -\mathbf{y}).
\end{equation} 

% We first state a lemma, which is proved in Appendix~\ref{app:propertiesofP}. 
We first state a lemma, which shows that the gradient estimation of {\cpml} is unbiased and variance bounded.
\begin{lemma}\label{lemma:sto_gradient}
Let $\mathbf{p}^{(t)} \triangleq \frac{1}{m} \overline{\bX}^{\top} \big( \bar{s}(\overline{\bX}, \overline{\bW}^{(t)}) - \mathbf{y} \big)$ denote the gradient computation using the quantized weights $\overline{\bW}^{(t)}$ in {\cpml}. Then, we have
\begin{itemize}[leftmargin=0.5cm]
    \item (Unbiasedness) Vector $\mathbf{p}^{(t)}$ is an asymptotically unbiased estimator of the true gradient.  $ \mathbb{E}[\mathbf{p}^{(t)}] = \nabla C (\bw^{(t)}) + \epsilon(r) $, and $\epsilon(r)\rightarrow \mathbf{0}$ as $r\rightarrow\infty$ where $r$ is the degree of the  polynomial in \eqref{eq:poly_approximation} and the expectation is taken with respect to the quantization errors,    
    \item (Variance bound) 
    $\mathbb{E}\big[\| \mathbf{p}^{(t)} - \mathbb{E}[\mathbf{p}^{(t)}] \|^{2}_{2} \big] \leq \frac{1}{2^{-2l_w}m^2}\| \bXq \|_{F}^{2} \triangleq \sigma^2$ where $\| \cdot \|_{2}$ and $\| \cdot \|_{F}$ are the  $l_2$ and Frobenius norms, respectively. 
\end{itemize}%\vspace{-0.1cm}
\end{lemma}

% \vspace{0.1cm}\emph{Proof.} 
% \input{appPropertiesOfP.tex}
\begin{proof}~\label{app:propertiesofP}
The proof of Lemma~\ref{lemma:sto_gradient} is presented in Appendix~A.
\end{proof}

% The proof of Lemma~\ref{lemma:sto_gradient} is presented in Appendix~\ref{app:sto_lemma_proof}.
% \subsection{Proof of Lemma~\ref{lemma:sto_gradient}}~\label{app:propertiesofP}
% \input{appPropertiesOfP.tex}

% We also need the following basic lemma, which is proved in Appendix~\ref{app:lipschitz}.
We also need the following basic lemma, which describes the $L$-Lipschitz property of the gradient of the cost function.
\begin{lemma}\label{lemma:lipschitz}
The gradient of the cost function from \eqref{cost} evaluated on the  quantized dataset $\overline{\bX}$ is $L$-Lipschitz with $L \triangleq \frac{1}{4}\| \overline{\bX} \|^{2}_{2}$, that is,  
$\| \nabla C(\mathbf{w}) - \nabla C(\mathbf{w}') \| \leq L\| \mathbf{w}-\mathbf{w}'\|$ for all $\mathbf{w}, \mathbf{w}' \in \mathbb{R}^d$. 
% \begin{equation}
%     \| \nabla C(\mathbf{w}) - \nabla C(\mathbf{w}') \| \leq L\| \mathbf{w}-\mathbf{w}'\|,
% \end{equation}
\end{lemma}

% \vspace{0.1cm}\emph{Proof.}
% \input{AppLipschitz.tex}

\begin{proof}~\label{app:lipschitz}
The proof of Lemma~\ref{lemma:lipschitz} is presented in Appendix B.
% The proof of Lemma~\ref{lemma:lipschitz} is presented in Appendix~\ref{app:lipschitz_proof}.
% \input{AppLipschitz.tex}
\end{proof}

We now state our main result for the theoretical performance  guarantees of {\cpml}.  
\begin{theorem}\label{thm:privacy} Consider the training of a logistic regression model  in a distributed system with $N$ workers using {\cpml} with the dataset $\mathbf{X} = (\mathbf{X}_{1}, \ldots, \mathbf{X}_{K})$, initial weight vector $\bw^{(0)}$,  and constant step size $\eta =1/L$ (where $L$ is defined in Lemma~\ref{lemma:lipschitz}). Then, {\cpml} guarantees,
\begin{itemize}
        \item (Convergence) $\mathbb{E}\big[ C\big( \frac{1}{J}\sum_{t=0}^{J}\bw^{(t)}\big)\big]-C(\bw^{*})\leq 
        \frac{{\| \bw^{(0)}-\bw^{*} \|}^2}{2\eta J}+\eta \sigma^2$ in $J$ iterations, where $\sigma^2$ is given in Lemma~\ref{lemma:sto_gradient},
        \item (Privacy) $\mathbf{X}$ remains information-theoretically private against any $T$ colluding workers, i.e., $I\big(\bX ; \widetilde{\mathbf{X}}_\mathcal{T}, \{\widetilde{\mathbf{W}}^{(t)}_\mathcal{T}\}_{t\in[J]}\big) = 0$ for any distribution on $\bX$ and any set $\mathcal{T}\subset [N]$ with $|\mathcal{T}| \leq T$,
    \end{itemize}
for any $N\geq (2r+1)(K+T-1) +1$, where $r$ is the degree of the polynomial from~\eqref{eq:poly_approximation}.
\end{theorem}
\begin{remark}\normalfont
Theorem~\ref{thm:privacy} reveals an important trade-off between privacy and parallelization in {\cpml}. Parameter $K$ reflects the amount of parallelization in {\cpml}, since the computation load at each worker node is proportional to $1/K$-th of the dataset. Parameter $T$ reflects the privacy threshold in {\cpml}. Theorem~\ref{thm:privacy} shows that, in a cluster with $N$ workers, we can achieve any  $K$ and $T$ as long as $N\geq (2r+1)(K+T-1) +1$. This condition  further implies that, as the number of workers $N$ increases, the parallelization ($K$) and privacy threshold ($T$) of {\cpml} can also increase linearly, leading to a scalable solution.
\end{remark} 
% \begin{remark}\normalfont
% Theorem~\ref{thm:privacy} applies also to the  simpler linear regression problem. The proof follows the same steps.
% \end{remark} 

\begin{remark}
There are two terms in the bound on the distance between the loss function to the optimum in the first equation of Theorem 1, i.e., $\mathbb{E}\big[ C\big( \frac{1}{J}\sum_{t=0}^{J}\bw^{(t)}\big)\big]-C(\bw^{*})\leq \frac{{\| \bw^{(0)}-\bw^{*} \|}^2}{2\eta J}+\eta \sigma^2$. When we use a constant learning rate $\eta=1/L$, the first term $\frac{{\| \bw^{(0)}-\bw^{*} \|}^2}{2\eta J}=\frac{L{\| \bw^{(0)}-\bw^{*} \|}^2}{2 J}$ goes to zero as the number of iterations $J$ increases, hence {\cpml} has the convergence rate of $O(1/J)$.
The second term $\eta \sigma^2 = \frac{\sigma^2}{L}$ is a residual error in the training as it does not go to zero as $J$ increases. 
By using an adaptive (decreasing) learning rate, this term can be made arbitrarily small. 
% If one uses an adaptive (decreasing) learning rate, this term will also vanish. 
\end{remark}

\begin{remark}
The convergence rate of {\cpml} is the same as that of conventional logistic regression. This follows from Theorem $1$ where the convergence rate of {\cpml} is found as $O(\frac{1}{J})$ where $J$ is the iteration index, which is the same as the convergence rate of conventional logistic regression, which follows from  \cite[Section 9.3]{boyd2004convex} and \cite[Section 7.1.1]{boyd2004convex}. 
\end{remark}

\begin{remark}\normalfont
Theorem~\ref{thm:privacy} applies also to (simpler) linear regression. The proof follows the same steps.
\end{remark} 

% \emph{Proof.} 
% We now state the proof of Theorem~\ref{thm:privacy}.

% \input{AppProofs.tex}

\begin{proof}
The proof of Theorem~\ref{thm:privacy} is presented in Appendix~C.
% The proof of Theorem~\ref{thm:privacy} is presented in Appendix~\ref{app:mainthm_proof}.
% We now state the proof of Theorem~\ref{thm:privacy}. 

% \input{AppProofs.tex}
% \input{AppProofs.tex}
\end{proof}

\vspace{-0.4cm}
\subsection{Complexity Analysis}
\begin{table}[t!]
\vspace{-0.3cm}
\small
\caption{Complexity summary of {\cpml}. }
\label{tbl:complexity_CPML}
\vspace{-0.5cm}
\begin{center}
% \begin{small}
% \begin{sc}
\begin{tabular}{ccc}
\toprule
 & Computation & Communication   \\
\midrule
 Master  & $O\big(\frac{mdN(K+T)}{K} + drJN(K+T)\big)$  & $O(\frac{mdN}{K} + drNJ)$  \\
 Worker  & $O(\frac{md^2}{K})$                          & $O(\frac{md}{K} + drJ)$  \\
\bottomrule
\end{tabular}
% \end{sc}
% \end{small}
\end{center}
\vspace{-0.5cm}
\end{table}

In this section, we analyze the asymptotic complexity of {\cpml} with respect to the number of workers $N$, parallelization parameter $K$, privacy parameter $T$, number of samples $m$, number of features $d$, and  number of iterations $J$.

\noindent {\bf Complexity Analysis of the Master Node:}
Computation cost of the master node can be broken into three parts: 1) encoding the dataset by using $\widetilde{\bX}_i=u(\alpha_i)$ from~\eqref{eq:lag1} for $i\in[N]$, 2) encoding the weight vector by using $\widetilde{\bW}_i^{(t)}=v(\alpha_i)$ from~\eqref{eq:lag2} for $i\in[N],t\in[J]$, and 3) decoding the gradient by recovering $h(\beta_i)$ in~\eqref{eq:beta} for $i\in[K]$.
For the first part, the encoded dataset $\widetilde{\mathbf{X}}_i$ ($i\in[N]$) from \eqref{eq:lag1} is a weighted sum of $K+T$ matrices where the size of each matrix is $\frac{m}{K}\times d$. 
The Lagrangian coefficients can be calculated offline since the sets of $\{\alpha_i\}_{i\in[N]}$ and $\{\beta_j\}_{j\in[K]}$ are public.
Each encoding requires $O\big(\frac{md(K+T)}{K}\big)$ multiplications and we must perform $N$ encodings, resulting in a total computational cost of $O\big(\frac{mdN(K+T)}{K}\big)$.
% As there are $N$ encoded dataset, this requires $O\big(\frac{mdN(K+T)}{K}\big)$ multiplications in total. For the second part, the encoded weight vector $\widetilde{\bW}_i^{(t)} (i\in[N],t\in[J])$ from~\eqref{eq:lag2} is also a weighted sum of $K+T$ matrices where the size of each matrix is $\frac{d}\times r$, which requires $O\big(drJN(K+T)\big)$ multiplications in total.
Decoding the gradient computations from \eqref{eq:decoding} can be performed via a weighted sum of $(2r+1)(K+T-1)+1=O(N)$ vectors where the size of each vector is $d$. Each decoding requires $O(dN)$ multiplications and we require $K$ decoded gradients, resulting in a total computational cost of $O(dJNK)$. 
% Note that the decoding overhead is negligible since $m$ is much larger than the other parameters. 
Communication cost of the master node to send the encoded dataset $\widetilde{\bX}_i$ and the encoded weight vector $\widetilde{\bW}_i^{(t)}$ to worker $i\in[N]$ is $O(\frac{mdN}{K})$ and $O(drNJ)$, respectively. 
Communication cost of the master to receive the local computation $f(\widetilde{\mathbf{X}}_i,\widetilde{\mathbf{W}}^{(t)}_i)$ from worker $i\in[N]$ for $t\in[J]$ is $O(dJN)$.

\noindent {\bf Complexity Analysis of the Workers:}
Computation cost of worker $i$ to compute $\widetilde{\mathbf{X}}_i^{\top}\widetilde{\mathbf{X}}_i$, the dominant part of the local computation $f(\widetilde{\mathbf{X}}_i,\widetilde{\mathbf{w}}^{(t)}_i)$ in \eqref{eq:compute}, is $O(\frac{md^2}{K})$.
This corresponds to $O(\frac{1}{K})^{th}$ of the computation cost of conventional logistic regression, which requires the computation of  $\mathbf{X}^{\top}s(\mathbf{X}\times \mathbf{w}^{(t)})$ in \eqref{grad}.
This is due to the fact that the size of the encoded dataset $\widetilde{\bX}_i$ and original dataset $\bX$ are $\frac{m}{K}\times d$ and $m\times d$, respectively.
Communication cost of worker $i$ to receive the encoded dataset $\widetilde{\bX}_i$ and the encoded weight vector $\widetilde{\bW}_i^{(t)}$ for $t\in[J]$ is $O(\frac{md}{K})$ and $O(drJ)$, respectively. Communication cost of worker $i$ to send the local computation $f(\widetilde{\mathbf{X}}_i,\widetilde{\mathbf{W}}^{(t)}_i)$ to the master for $t\in[J]$ is $O(dJ)$.

We summarize the asymptotic complexity of {\cpml} in Table~\ref{tbl:complexity_CPML}.
\vspace{-0.2cm}

\section{Experiments}\label{section:experiments}
We now experimentally demonstrate the performance of {\cpml} compared to conventional MPC baselines. Our focus is on training a logistic regression model for image classification, while the computation load is distributed to  multiple machines on the Amazon EC2 Cloud Platform.

% {\bf Setup.}
\noindent{\bf Experiment setup.}
% We train the logistic regression model from~\eqref{cost} for binary image classification on the MNIST \cite{lecun2010mnist} and CIFAR-10 \cite{krizhevsky2009learning} datasets to experimentally examine two things: the accuracy of {\cpml} and the performance gain in terms of training time. The size of the MNIST and CIFAR-10 datasets are $(m,d)=(12396,784)$ and $(9019,3073)$, respectively.
We train the logistic regression model from~\eqref{cost} for binary image classification on the CIFAR-10 \cite{krizhevsky2009learning} and GISETTE \cite{NIPS2004_2728} datasets to experimentally examine two things: the accuracy of {\cpml} and the performance gain in terms of training time. The size of the CIFAR-10 and GISETTE datasets are $(m,d)=(9019,3073)$\footnote{We select images with the label of "plane" and "car", and the number of these images in $50000$ training samples is $9019$. For the number of features, we added a bias term, hence, we have $3072+1=3073$ features.} and $(6000,5000)$, respectively. 
% \footnote{To have a larger dataset we duplicate the MNIST dataset.}. 
We implement  the communication phase using the {\tt MPI4Py}~\cite{dalcin2011parallel} message passing interface on {\tt Python}. 
Computations are performed in a distributed manner on Amazon EC2 clusters using  \texttt{m3.xlarge} machine instances. 

% We then compare {\cpml} with two MPC-based benchmarks that we apply to our problem. In particular, we implement two MPC constructions. The first one is based on the well-known BGW protocol~\cite{ben1988completeness}, whereas the second one is a more recent protocol from \cite{beerliova2008perfectly, damgaard2007scalable} that trade-offs offline calculations for a more efficient implementation. Both baselines utilize Shamir's secret sharing scheme~\cite{shamir1979share} where the dataset is secret shared among the $N$ workers who proceed with a multiround protocol to compute the gradient. 
% We further incorporate our quantization and approximation techniques as conventional MPC protocols are also bound to arithmetic operations over a finite field. The implementation details are provided in Appendix~\ref{app:BGW}. 

We then compare {\cpml} with two MPC-based benchmarks that we apply to our problem. 
In particular, we implement two MPC constructions. 
The first one is based on the well-known BGW protocol~\cite{ben1988completeness}, whereas the second one is a more recent protocol from \cite{beerliova2008perfectly, damgaard2007scalable} that trade-offs offline calculations for a more efficient implementation. Our choice of these MPC benchmarks is due to their ability to be applied to a  large number of workers. While several more recent works exist that have developed MPC-based training protocols with information-theoretic privacy guarantees, their constructions are limited to three or four parties~\cite{mohassel2017secureml, cryptoeprint:2018:442,mohassel2018aby}. For instance, \cite{mohassel2017secureml} is a two-party protocol that requires two non-colluding workers. 

Both baselines utilize Shamir's secret sharing scheme~\cite{shamir1979share} where the dataset is secret shared among the $N$ workers. 
For the (quantized) dataset $\bXq$,
% $\overline{\bX}=[\overline{\bX}^{\top}_1 \ldots \overline{\bX}^{\top}_K]^{\top}$, where $\overline{\bX}_i \in \mathbb{F}^{\frac{m}{K}\times d}_{p}$, 
% this is achieved by creating a random polynomial $\mathbf{P}_i(z) = \bXq_i + z \mathbf{Z}_{i1}+ \ldots + z^T \mathbf{Z}_{iT}$  for each $i\in [K]$,  where $\mathbf{Z}_{ij}$ for $j\in [T]$ are i.i.d. uniformly distributed random matrices.
this is achieved by creating a random polynomial $\mathbf{P}(z) = \bXq + z \mathbf{Z}_{1}+ \ldots + z^T \mathbf{Z}_{T}$, where $\mathbf{Z}_{j}$ for $j\in [T]$ are i.i.d. uniformly distributed random matrices.
% Then, each worker is assigned a secret share of the dataset using this polynomial. 
This guarantees privacy against $\lfloor \frac{N-1}{2}\rfloor$ colluding workers \cite{ben1988completeness, beerliova2008perfectly, damgaard2007scalable}, 
% which is larger than the privacy threshold of {\cpml}, 
but requires a computation load at each worker that is as large as processing the whole dataset 
at a single worker, leading to slow training. Hence, in order to provide a fair comparison with {\cpml},
% specifically, each worker receives a share for every $i\in [K]$. Hence, the total amount of data stored at each worker is equal to the size of the whole dataset $\overline{\bX}$. A similar polynomial is created for secret sharing the quantized weights $\overline{\mathbf{W}}^{(t)}$.
% The workers then proceed with a multiround protocol to compute the gradient. 
% We implement the communication phase using the  {\tt MPI4Py} message passing interface.   
% {\color{red} The time spent during communication is included in the reported computation time. } 
% We further incorporate our quantization and approximation techniques as conventional MPC protocols are also bound to arithmetic operations over a finite field. 
% The system parameters used for quantization and polynomial approximation are selected to be the same as the ones used for {\cpml}. 
we optimize (speed up) the benchmark protocols by partitioning the users into subgroups of size $2T+1$. Then, we let each group compute the gradient over the partitioned dataset $\overline{\bX}'_i \in \mathbb{F}^{\frac{m}{G}\times d}_{p}$, where $\overline{\bX}=[\overline{\bX}'^{\top}_1 \ldots \overline{\bX}'^{\top}_G]^{\top}$ and $G$ is the number of subgroups.
For group $i\in[G]$, each worker receives a share of the partitioned dataset by using a random polynomial $\mathbf{P_i}(z) = \bXq'_i + z \mathbf{Z}_{i1}+ \ldots + z^T \mathbf{Z}_{iT}$, where $\mathbf{Z}_{ij}$ for $j\in [T]$ and $i\in[G]$ are i.i.d. uniformly distributed random matrices. 
Workers then proceed with a multiround protocol to compute the sub-gradient. 
% We implement the communication phase using the  {\tt MPI4Py} message passing interface.   
We further incorporate our quantization and approximation techniques in our benchmark implementations as conventional MPC protocols are also bound to arithmetic operations over a finite field. 
In our experiments, 
% for a fair comparison with {\cpml}, 
we set $G=3$, hence the total amount of data stored at each worker is equal to one third of the size of the dataset $\overline{\bX}$, which significantly reduces the total training time of the two benchmarks, while providing a privacy threshold of $T=\lfloor \frac{N-3}{6}\rfloor$. 
The implementation details of the MPC operations are provided in Appendix~D.
% The implementation details of the MPC operations are provided in Appendix~\ref{app:BGW}.

\noindent{\bf {\cpml} parameters.} There are several system parameters in {\cpml} that should be set. 
Given that we have a $64$-bit implementation, we  select the field size to be $p=2^{25}-37$, which is the largest prime with $25$ bits to avoid an overflow on intermediate multiplications.  
% Given that we have a $64$-bit implementation, we  select the field size to be $p=67108859$, which is the largest prime with $26$ bits to avoid an overflow on intermediate multiplications
We then optimize the quantization parameters, $l_x$ in \eqref{eq:X_q} and $l_w$ in \eqref{eq:vector}, by taking into account the trade-off between the rounding and overflow error. In particular, we choose $(l_x,l_w)=(2,6)$ and $(2,5)$ for the CIFAR-10 and GISETTE datasets, respectively. 
% In particular, we choose $(l_x,l_w)=(2,4)$ and $(2,7)$ for the MNIST and CIFAR-10 datasets, respectively. 
We also need to set the parameter $r$, the degree of the polynomial for approximating the sigmoid function. We consider both $r=1$ and $r=2$ and as shown later empirically we observe that the degree one approximation achieves good accuracy. We finally need to select $T$ (privacy threshold) and $K$ (amount of parallelization) in {\cpml}. As stated in Theorem~\ref{thm:privacy}, these parameters should satisfy $N\geq (2r+1)(K+T-1) +1$. Given our choice of $r=1$, we consider two cases:
\begin{itemize}[leftmargin=0.5cm]
\item \textbf{Case 1 (maximum parallelization).}
All resources allocated for parallelization (faster  training) by setting $K= \lfloor \frac{N\!-\!1}{3} \rfloor$, $T=1$,  

% \item \textbf{Setup 2 (equal parallelization \& privacy).} Resources split equally between parallelization \& privacy, i.e., $K=T=\lfloor \frac{N+2}{6} \rfloor$.
\item \textbf{Case 2 (equal parallelization \& privacy).} Resources split almost equally between parallelization \& privacy, i.e., $T=\lfloor \frac{N-3}{6} \rfloor, K=\lfloor \frac{N+2}{3} \rfloor-T$.  
\end{itemize}

\begin{figure}%
    \centering
    % \subfigure[MNIST (for accuracy $95.44\%$ with $25$ iterations)]{%
    % \label{fig:first}%
    % \includegraphics[width=0.82\linewidth]{figures/trainingtime_trustmaster_smalldataset.png}}% 
    % \\
    % \vspace{-0.3cm}
    \subfigure[CIFAR-10 (for accuracy $81.35\%$ with $50$ iterations)]{%
    \label{fig:first}%
    \includegraphics[width=0.8\linewidth]{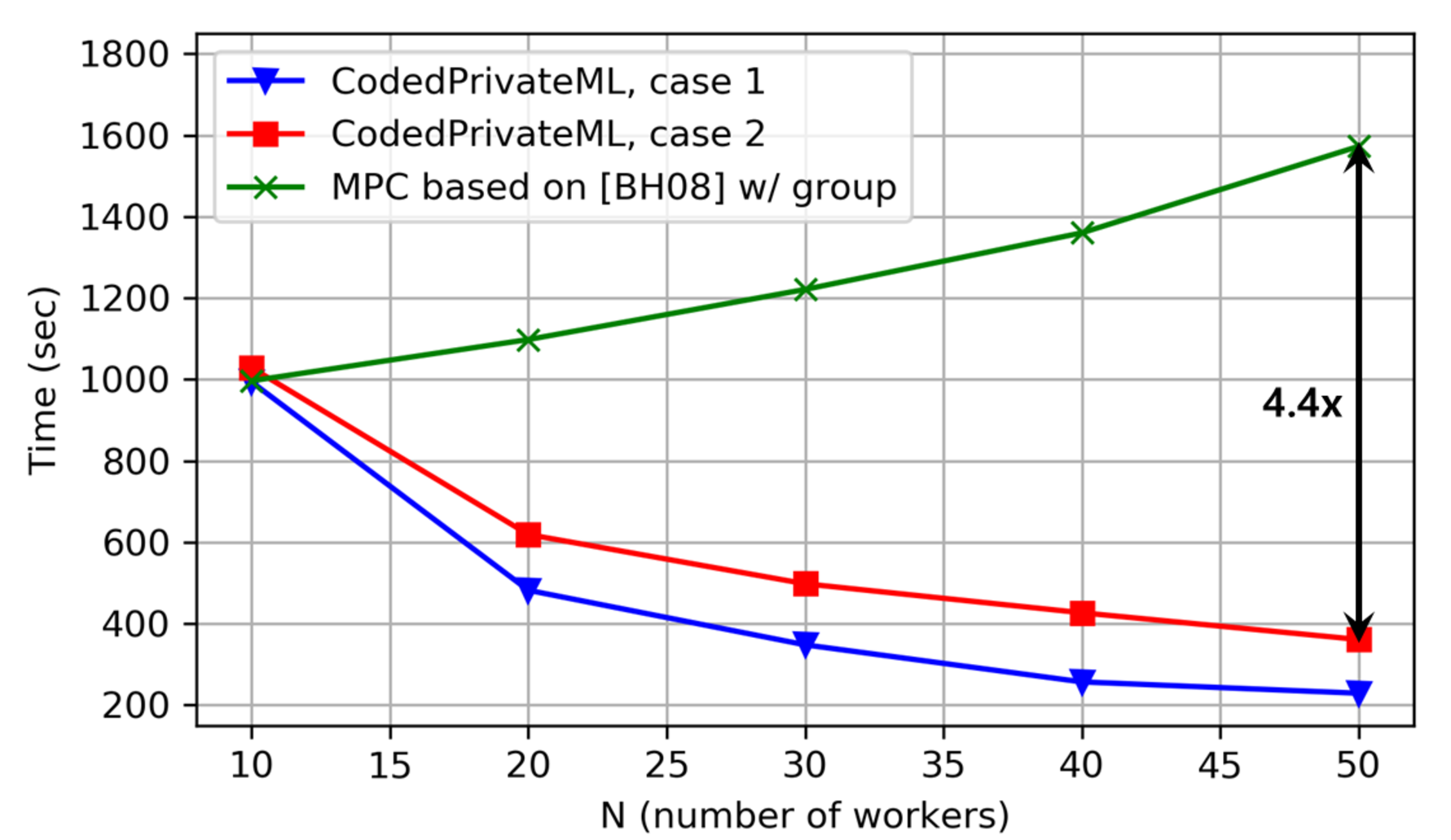}}%
    \\
    \vspace{-0.3cm}
    \subfigure[GISETTE (for accuracy $97.50\%$ with $50$ iterations)]{%
    \label{fig:second}%
    \includegraphics[width=0.8\linewidth]{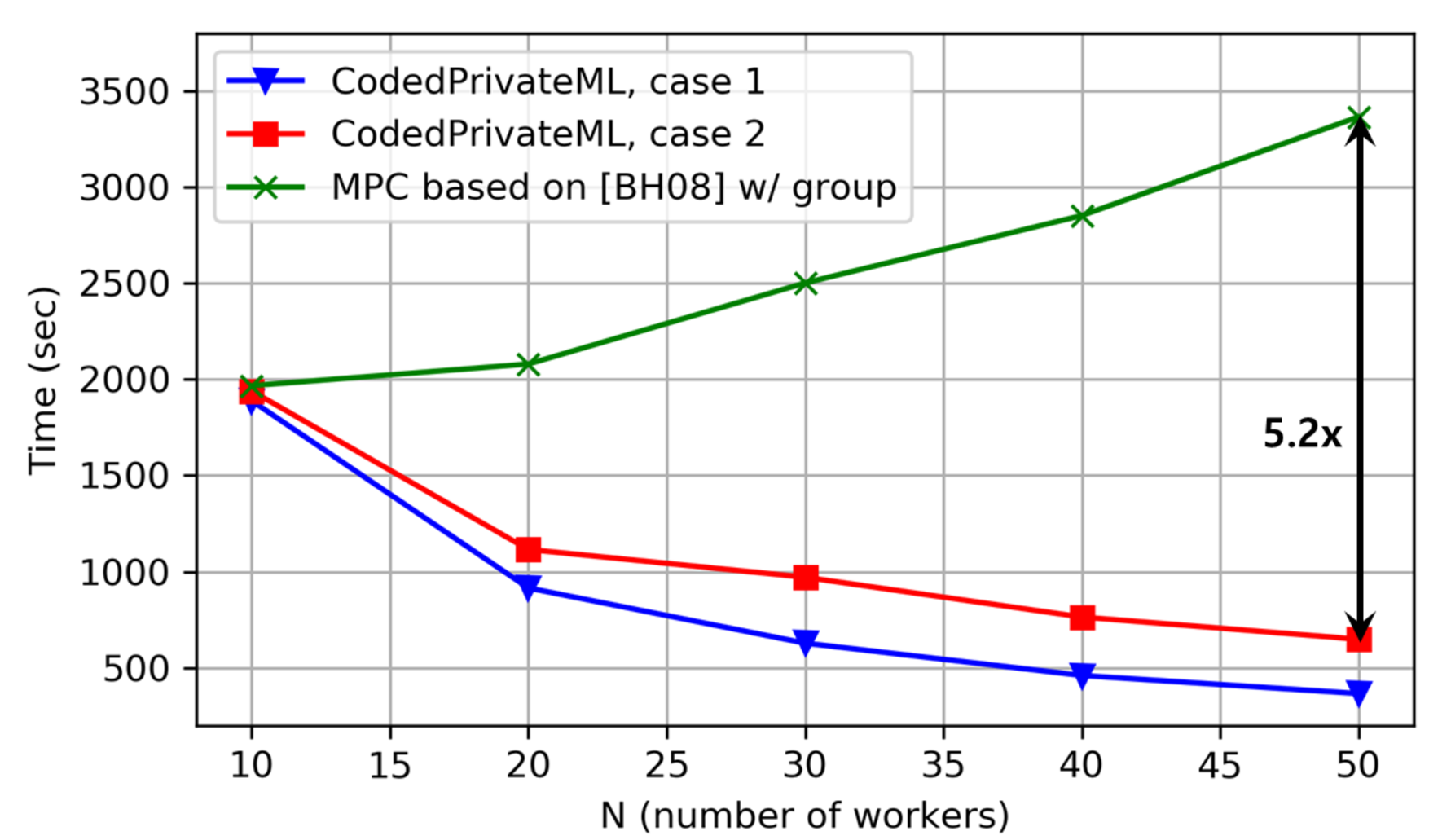}}%
% \vspace{-0.4cm}\caption{Performance gain of {\cpml} over the MPC baselines ([BGW88] from \cite{ben1988completeness} and [BH08] from \cite{beerliova2008perfectly}). The plot shows the total training time for different number of workers $N$.
\vspace{-0.1cm}\caption{Performance gain of {\cpml} over the MPC baseline ([BH08] from \cite{beerliova2008perfectly}). The plot shows the total training time for different number of workers $N$. 
}
\label{fig:trainingtimeSec5}
\vspace{-0.2cm}
\end{figure}

\vspace{0.1cm}
\noindent{\bf Training time.}
Initially, we measure the training time while increasing the number of workers $N$ gradually. Our results are demonstrated in Figure~\ref{fig:trainingtimeSec5}, which shows the comparison of  {\cpml} with the [BH08] protocol from \cite{beerliova2008perfectly}, as we have found it to be the faster of the two benchmarks. In particular, we make the following observations.\footnote{For $N=10$, all schemes have similar performance because the total amount of data stored at each worker is one third of the size of whole dataset ($K=3$ for {\cpml} and $G=3$ for the benchmark).}

% \footnote{
% {\color{red}For $N=5$, all schemes have similar performance as they use the same system parameters, $K=T=1$.}
% }

\begin{table}[t]
% \footnotesize
\small
 \caption{(CIFAR-10) Breakdown of total runtime for $N=50$.\vspace{-0.2cm}}
\label{table:CIFAR_N50}
  \centering
  \begin{tabular}{ l  c  c  c  c }
    \toprule
    % Protocol & \hspace{-0.0cm}Enc. time (s) & Comm. time (s)& Comp. time (s)& Total time (s) \\
    Protocol          & \hspace{-0.3cm}Enc.    & \hspace{-0.2cm}Comm.   & \hspace{-0.2cm}Comp.   & \hspace{-0.2cm}Total  \\
                      & \hspace{-0.3cm}time (s)& \hspace{-0.2cm}time (s)& \hspace{-0.2cm}time (s)& \hspace{-0.2cm}time (s) \\
     \midrule
    MPC using [BGW88] &  \hspace{-0.3cm}202.78 & \hspace{-0.2cm}31.02 & \hspace{-0.2cm}7892.42 & \hspace{-0.2cm}8127.07 \\ 
    MPC using [BH08]  &  \hspace{-0.3cm}201.08 & \hspace{-0.2cm}30.25 & \hspace{-0.2cm}1326.03 & \hspace{-0.2cm}1572.34 \\ 
    {\cpml} (Case 1)  &  \hspace{-0.3cm}59.93  & \hspace{-0.2cm}4.76  & \hspace{-0.2cm}141.72  & \hspace{-0.2cm}229.07  \\ 
    {\cpml} (Case 2)  &  \hspace{-0.3cm}91.53  & \hspace{-0.2cm}8.30  & \hspace{-0.2cm}235.18  & \hspace{-0.2cm}361.08  \\ 
    % {\cpml} (case3) & 826.10 & 49.74 & 812.41  & 1711.04 \\ \hline
    \bottomrule
  \end{tabular}
\vspace{-0.5cm}
\end{table}

\begin{figure}[t]%
    \centering
    % \subfigure[MNIST dataset, binary classification between digits 3 and 7 (using $12396$ samples for the training set and $2038$ samples for the test set).]{%
    % % \vspace{-0.7cm}
    % \label{fig:accuracySec5_MNIST}%
    % \includegraphics[width=0.85\linewidth]{figures/accuracy_trustmaster_MNIST.png}}%
    % \\
    % \vspace{-0.3cm}
    \subfigure[CIFAR-10 dataset, binary classification between \emph{car} and \emph{plane} images (using $9019$ samples for the training set and $2000$ samples for the test set).]{%
    % \vspace{-0.7cm}
    \label{fig:accuracySec5_CIFAR}%
    \includegraphics[width=0.8\linewidth]{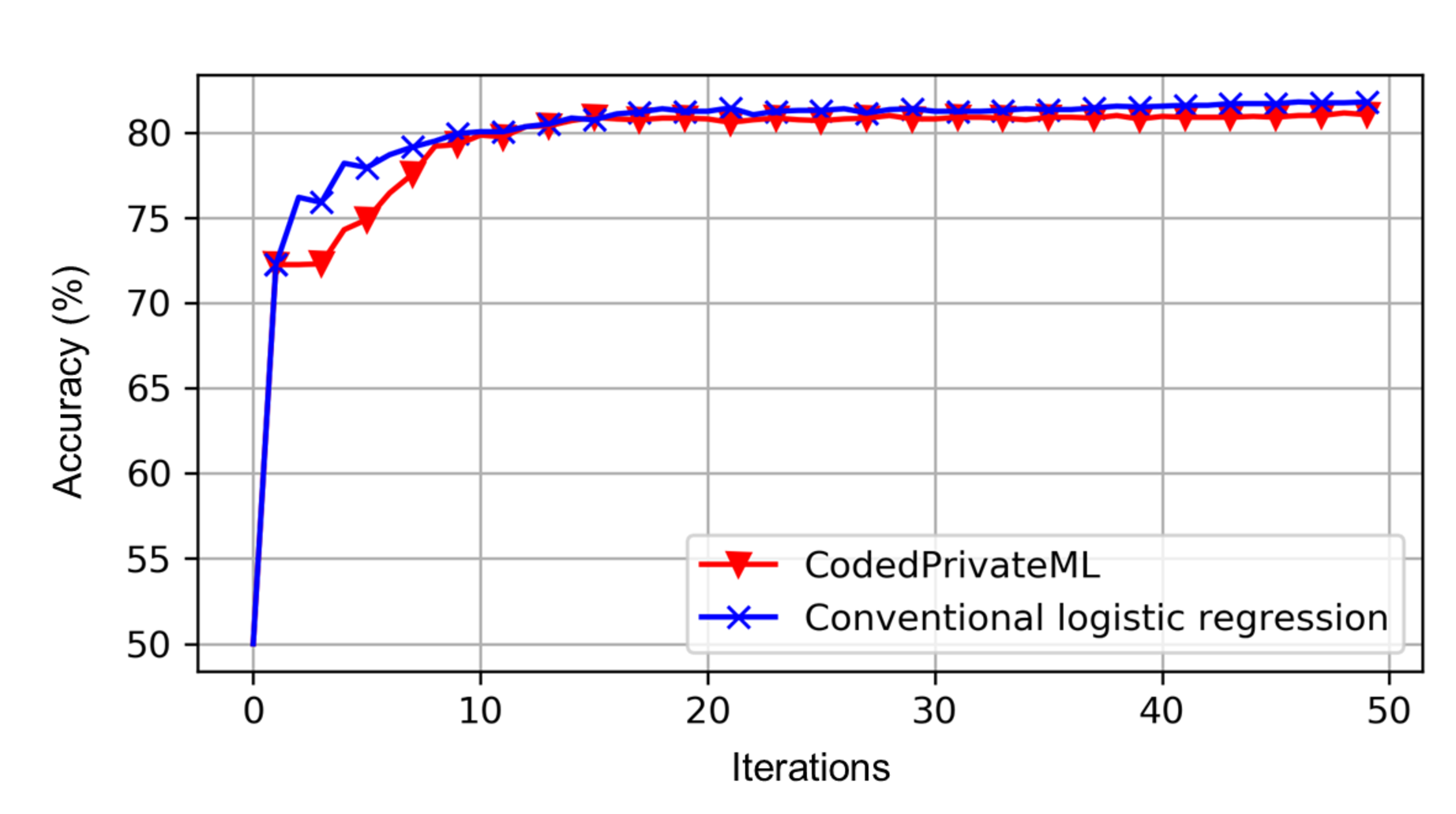}}%
    \\
    \vspace{-0.3cm}
    \subfigure[GISETTE dataset, binary classification between the images of digits 4 and 9 (using $6000$ samples for the training set and $1000$ samples for the test set).]{%
    % \vspace{-0.7cm}
    \label{fig:accuracySec5_GISETTE}%
    \includegraphics[width=0.8\linewidth]{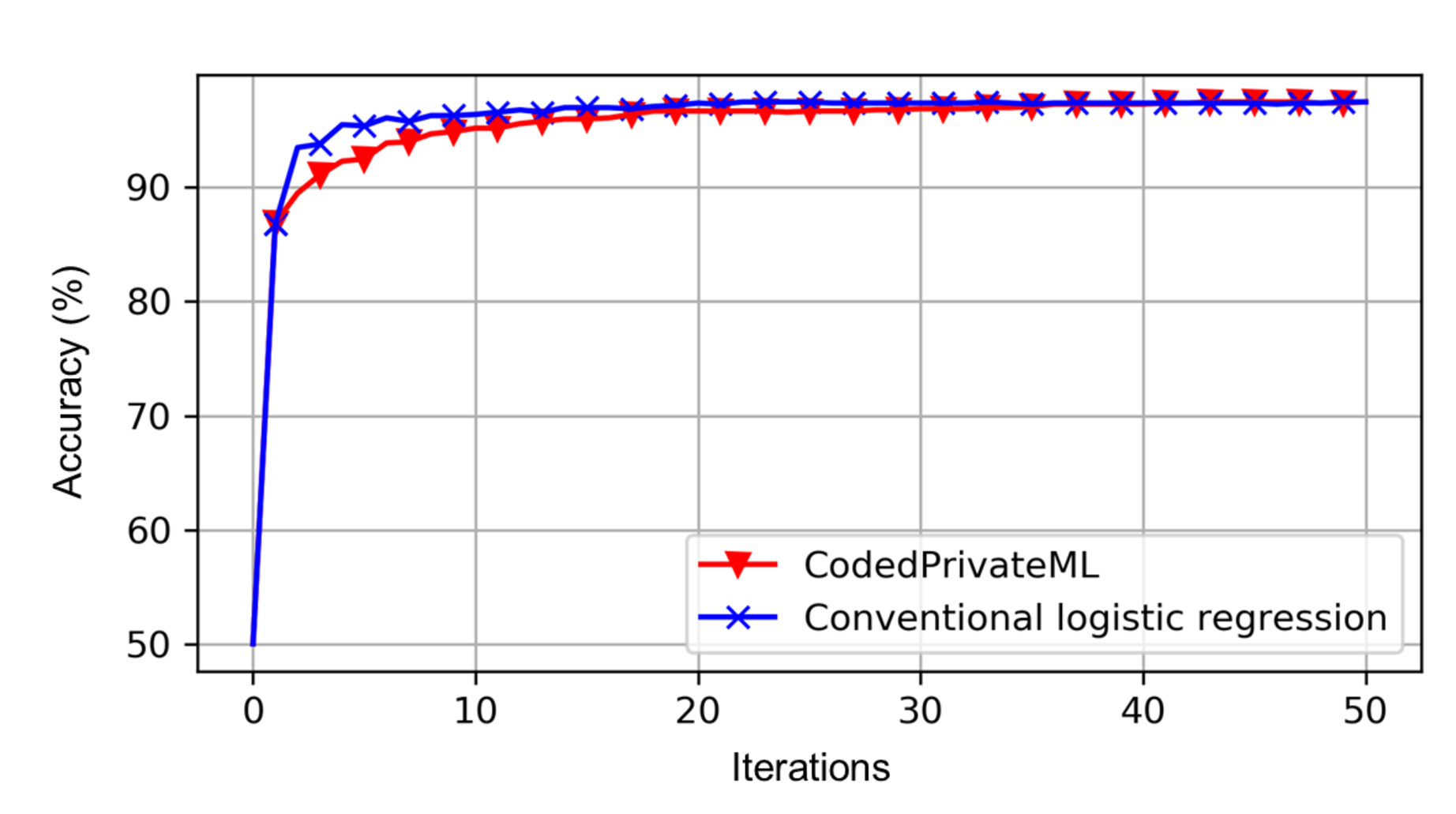}}%
\vspace{-0.1cm}\caption{Comparison of the accuracy of {\cpml} (demonstrated for Case 2 and $N=50$ workers) vs  conventional logistic regression that uses the sigmoid function without quantization.}
\label{fig:accuracySec5}
\vspace{-0.5cm}\end{figure}

\begin{itemize}[leftmargin=0.3cm]
\item {\cpml} provides substantial speedup over the MPC baselines, in particular, up to $4.4\times$ and $5.2\times$ with the CIFAR-10 and GISETTE datasets, respectively, while providing the same privacy threshold as the benchmarks ($T=\lfloor \frac{N-3}{6} \rfloor$ for Case 2).
% \item {\cpml} provides substantial speedup over the MPC baselines, in particular, up to $4.4\times$ and $5.2\times$ with the CIFAR-10 and GISETTE datasets, respectively, while Setup 2 of {\cpml} and benchmark with grouping $G=3$ have the same privacy threshold $T=\lfloor \frac{N-3}{6} \rfloor$.
% Tables~\ref{table:CIFAR_N40}-\ref{table:CIFAR_N10} demonstrate the breakdown of the total runtime with CIFAR-10 dataset for $N=40$, $N=25$, and $N=10$ workers, respectively.
Table~\ref{table:CIFAR_N50} demonstrates the breakdown of the total runtime with the CIFAR-10 dataset for $N=50$ workers.
In this scenario, {\cpml} provides significant improvement in all three categories of dataset encoding and secret sharing; communication time between the workers and the master; and the computation time. 
Main reason for this is that, in the MPC baselines, the size of the data processed at each worker is one third of the original dataset, while in {\cpml} it is $1/K$-th of the dataset. This reduces the computational overhead of each worker while computing matrix multiplications as well as the communication overhead between the master and workers. 
% In addition, the MPC baseline requires additional communication between the workers to execute a degree reduction phase for every multiplication gate. 
% This provides a larger parallelization gain for {\cpml}. The other reason is the communication complexity of MPC-based scheme. 
We also observe that a higher amount of speedup is achieved as the dimension of the  dataset  becomes larger (CIFAR-10 vs. GISETTE datasets), suggesting {\cpml} to be well-suited for data-intensive training tasks where parallelization is essential.

% Results for additional scenarios are provided in  Appendix~\ref{app:Exp}.
\item The total runtime of {\cpml} decreases as the number of workers increases. This is again due to the parallelization gain of {\cpml} (i.e., increasing $K$ while $N$ increases). 
This is not achievable in conventional MPC baselines, since the size of data processed at each worker is constant for all $N$.
% This parallelization gain is not achievable in conventional MPC baselines, since the whole computation has to be repeated by all players who take part in the MPC. 
% We should, however, point out that the MPC baselines can attain a higher privacy threshold ($T=N/2-1$), while {\cpml} can achieve $T=\lfloor \frac{N+2}{6} \rfloor$ (Setup 2). In other words, to guarantee the same amount of privacy, {\cpml} requires more workers. 

\item Increasing $N$ in {\cpml} has two major impacts on the total training time. The first one is reducing the computation load per worker, as each new worker can be used to increase the parameter $K$. This in turn reduces the computation load per worker as the amount of work done by each worker is scaled with respect to $1/K$. The second one is that increasing the number of workers increases the encoding time at the master node. 
% For small datasets, i.e., when the computation load at each worker is very small, the gain from increasing the number of workers beyond a certain point may be minimal and the system may saturate. 
Hence, the gain from increasing the number of workers beyond a certain point may be minimal and the system may saturate. 
In those cases, increasing the number of workers cannot further reduce the training time, as the computation will be dominated by the encoding overhead.
% In the small dataset regime using the CIFAR-10 dataset (Fig. \ref{fig:first}), we can observe that after about $N=30$ workers, the cost of encoding time becomes dominant over the gains obtained from Lagrange coding, and as a result the total training time starts to increase.
% On the other hand, in a larger dataset regime using the GISETTE dataset (Fig. \ref{fig:second}), we observed that the training time reduces as the number of clients increases until $N=50$.

\item 
{\cpml} provides up to $22.5\times$ speedup over the BGW protocol~\cite{ben1988completeness}, as shown in Table~\ref{table:CIFAR_N50} for the CIFAR-10 dataset with $N=50$ workers. 
This is due to the fact that BGW requires additional communication between the workers to execute a degree reduction phase for every multiplication operation. 
\end{itemize}

\noindent{\bf Accuracy.} 
We also examine the accuracy and convergence of {\cpml}.  
Figure~\ref{fig:accuracySec5_CIFAR} illustrates the test accuracy of the binary classification problem between \emph{plane} and \emph{car} images for the CIFAR-10 dataset. With 50 iterations, the accuracy of {\cpml} with degree one polynomial approximation and conventional logistic regression are $81.35\%$ and $81.75\%$, respectively. 
Figure~\ref{fig:accuracySec5_GISETTE} shows the test accuracy for binary classification between digits 4 and 9 for the GISETTE dataset. With 50 iterations, the accuracy of {\cpml} with degree one polynomial approximation and conventional logistic regression has the same value of $97.5\%$. 
Hence,  {\cpml} has comparable accuracy to conventional logistic regression while being privacy preserving.

%%%%%%%%%%

% We next present the breakdown of the total runtime for training on the MNIST dataset by varying the number of workers. 
% Tables~\ref{table:MNIST_N10} and \ref{table:MNIST_N25} demonstrate the corresponding results for $N=10$ and $N=25$ workers, respectively.  
% One can note that, in all scenarios, {\cpml} provides significant improvement in all three categories of dataset encoding and secret sharing; communication time between the workers and the master; and computation time. 
%Upon investigating the corresponding tables, we observe that the performance gain of {\cpml} over the MPC-based schemes increases as the number of workers increase. 

%%% CIFAR 10

\begin{figure}[t]
% \vskip 0.2in
\begin{center}
% \centerline{\includegraphics[width=0.6\columnwidth]{figures/cross_entropy_larger_dataset.png}}
\centerline{
    \includegraphics[width=0.8\linewidth]{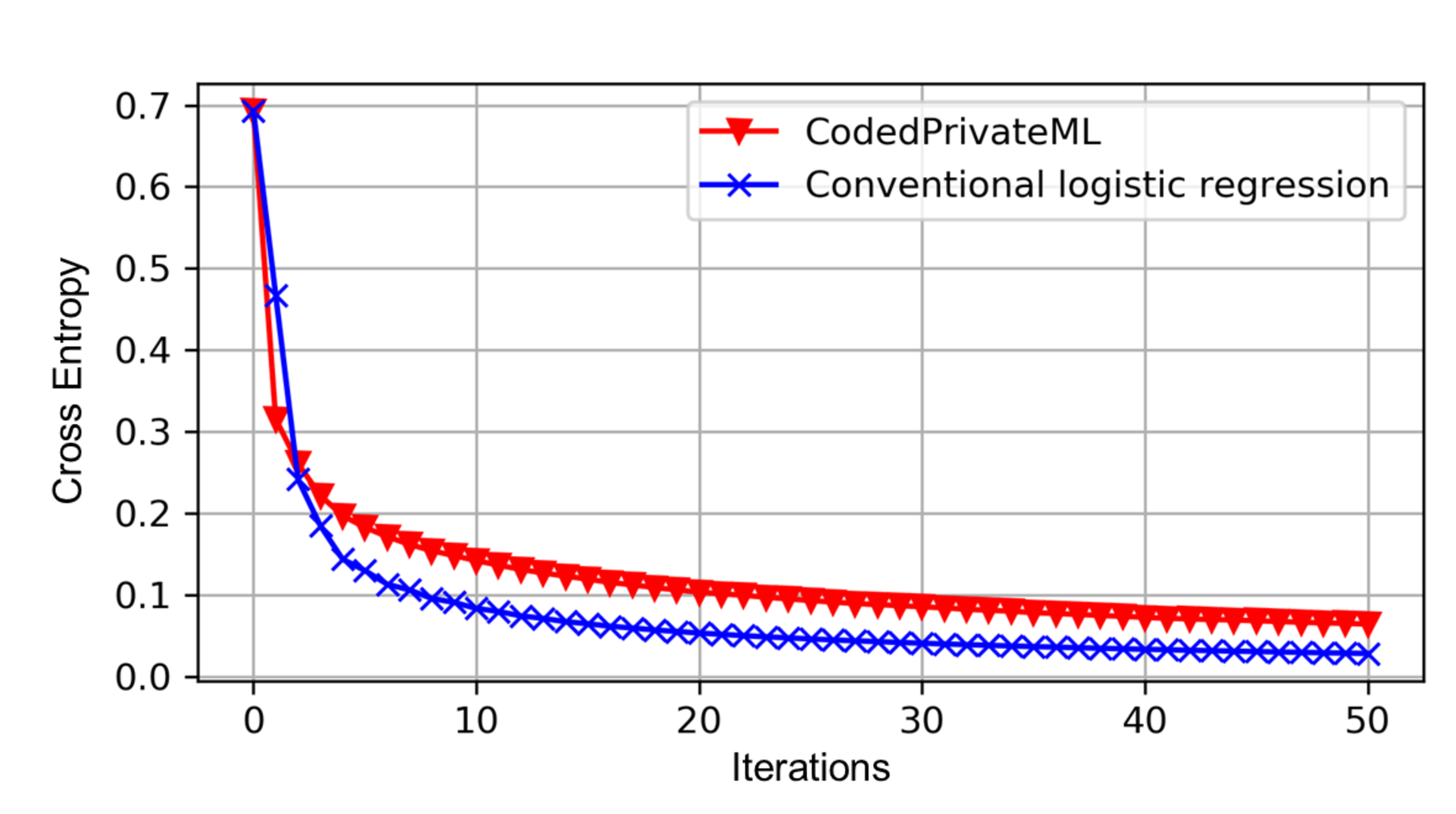}}
    \vspace{-0.3cm}
    \caption{Convergence of {\cpml} (demonstrated for Case 2 and $N=50$ workers) vs  conventional logistic regression (using the sigmoid function without  polynomial approximation or quantization). %(demonstrated for Setup 2 and $N=40$, i.e., $(N,K,T)=(40,7,7)$) 
% Some workers may straggle or fail to return the computations, due to slow processing speed or system failures.
}
\label{fig:cross_entropy}
\end{center}
\vspace{-0.5cm}
\end{figure}

% \subsubsection{Convergence of {\cpml}} 
% We also experimentally analyze the convergence behaviour of {\cpml}.
Figure~\ref{fig:cross_entropy} presents the cross entropy loss for {\cpml} versus the conventional logistic regression model for the GISETTE dataset. The latter setup  uses the sigmoid function and no polynomial approximation, in addition, no quantization is applied to the dataset or the weight vectors. We observe that {\cpml} achieves convergence with comparable rate to conventional logistic regression, while being privacy preserving. 
% Thus, {\cpml} achieves convergence with comparable rate to conventional logistic regression, while being privacy preserving. 
% Therefore, {\cpml} guarantees almost the same convergence rate, while being privacy preserving.

\vspace{-0.2cm}
\section{Conclusion and Discussion}\label{section:conclusion}
In this paper, we considered a distributed training scenario in which a data-owner wants to train a logistic regression model by off-loading the computationally-intensive gradient computations to multiple workers, while preserving the privacy of the dataset. We proposed a privacy-preserving training framework, {\cpml}, that distributes the computation load effectively across multiple workers, and reduces the per-worker computation load as more and more workers become available. We demonstrated the theoretical convergence guarantees and the fundamental trade-offs of our framework, in terms of the number of workers, privacy protection, and scalability. Our experiment results demonstrate significant speed-up in the training time compared to conventional baseline protocols.  

% \subsection{Discussion}
This work focuses on a logistic regression model mainly with the goal of demonstrating how {\cpml} can be utilized to scale and speed up logistic regression training under privacy and convergence guarantees, which is a first step towards more complex models.
To the best of our knowledge, even for this setup, no other system has been able to efficiently scale beyond $3-4$ workers while achieving information-theoretic privacy.
Our work is the first privacy-preserving machine learning approach that reduces the communication and computation load per worker as the number of workers increases, which we hope will open up further research. 
Future directions include extending {\cpml} to deeper neural networks by leveraging an MPC-friendly (i.e., polynomial) activation function or extending {\cpml} to collaborative learning setting such as~\cite{so2020scalable}.
% such as the one proposed in \cite{mohassel2017secureml}. 
% {\cpml} can also be extended to deeper neural network as an MPC-friendly (i.e., polynomial) activation function is proposed in \cite{mohassel2017secureml,gilad2016cryptonets,hesamifard2017cryptodl,chabanne2017privacy} and shows that the accuracy of the resulting models is very close to those trained using the original functions. 
% We expect to achieve a similar performance gain even in those setups, since {\cpml} can similarly be leveraged to efficiently parallelize the MPC computations. 
% However, the error rate could increase as the number of nonlinear layer being approximated with low degree polynomial is large, which we can find from deeper neural network \cite{gilad2016cryptonets}.
% In addition, piece-wise linear model to approximate the ReLU function \cite{mohassel2017secureml} or max-pooling layer requires an expensive multiparty secure comparison protocol \cite{kerschbaum2009performance}.
% The multiparty secure comparison protocol, which is based on secret shares, has a round complexity of $O(\log{N})$, communication complexity of $O(N^2 \log{N})$, and computation complexity of $O(N \log{N})$ \cite{kerschbaum2009performance}.
% This $O(N^2 \log{N})$ communication complexity could be the major bottleneck to scale deep neural network beyond $3-4$ workers.
% Addressing these issues to extend {\cpml} to the deep neural network is challenging and would be a very interesting future direction.

In this paper, in order to provide information-theoretic privacy, we utilize quantization to convert the dataset and model to the finite field $\mathbb{F}_p$. Doing so has two inherent challenges: 1) determining a proper value for $p$ and 2) potential performance degradation caused by quantization or overflow error.
This has inspired a new line of works, such as \emph{analog coded computing} \cite{soleymani2020analog,soleymani2020privacy}, which uses floating-point numbers instead of fixed-point numbers to represent the finite field and provides a fundamental trade-off between the accuracy and privacy level. Leveraging such techniques to address these challenges is another interesting future direction. 

\vspace{-0.25cm}
\section*{Acknowledgement}
\vspace{-0.1cm}
Authors would like to thank helpful comments and discussions with Dr. Payman Mohassel on this problem.
\vspace{-0.2cm}

\bibliographystyle{unsrt}
\bibliography{Main.bbl}

\clearpage

\onecolumn

\appendix

% \subsection{Supplementary Materials}

\subsection{Proof of Lemma~\ref{lemma:sto_gradient}}\label{app:sto_lemma_proof}
(Unbiasedness) Given  $\overline{\bX}$, we have
\begin{equation}\label{eq:unb}
\mathbb{E}[\mathbf{p}^{(t)}] = \frac{1}{m}\overline{\bX}^{\top}\Big( \mathbb{E}\big[\bar{s}(\overline{\bX}, \overline{\bW}^{(t)})\big] - \mathbf{y} \Big) = \frac{1}{m} \overline{\bX}^{\top} \Big( \hat{s}(\overline{\bX} \times {\bw}^{(t)}) - \mathbf{y} \Big), 
\end{equation}
where \eqref{eq:unb} follows from the unbiasedness of the quantization strategy, $E[\bar{s}(\overline{\mathbf{X}}, \overline{\mathbf{W}}^{(t)}) ] =\hat{s}(\overline{\mathbf{X}}\times \mathbf{w}^{(t)})$, and expectation is taken with respect to the quantization noise at iteration $t$. Then, we obtain
\begin{equation}\label{eq:delta}
    \mathbb{E}[\mathbf{p}^{(t)}] - \nabla C(\bw^{(t)}) =\frac{1}{m}\bXq^{\top} \big( \hat{s}(\overline{\bX}\times {\bw}^{(t)}) - s(\overline{\bX}\times {\bw}^{(t)}) \big).  
\end{equation}
Assume $\bw^{(t)}$ is constrained such that $\| \bw^{(t)} \| \leq B$ for some real value $B\in \mathbb{R}$ \cite{ZipML_arXiv,zhou2018empirical}. 
Then, from the Weierstrass approximation theorem~\cite{brinkhuis2011optimization,pinkus2000weierstrass}, for every $\epsilon>0$, there exists a polynomial that approximates the sigmoid arbitrarily well, i.e.,  $|\hat{s}(x)-s(x)|\leq\epsilon$ for all $x$ in the constrained interval.  
% The domain for the sigmoid function is bounded as $\|\bw\|\leq R$. 
Therefore, given $\bXq$, there exists a polynomial making the norm of~\eqref{eq:delta} arbitrarily small.   

% $\mathbf{\delta}(r)$

(Variance bound) The variance of $\mathbf{p}^{(t)}$ satisfies,
% For simplicity, we omit the iteration index, $t$, for this proof.
\begin{align}
\mathbb{E}\big[\| \mathbf{p}^{(t)} - \mathbb{E}[\mathbf{p}^{(t)}] \|^{2}_{2} \big]   
&= \frac{1}{m^2}\mathbb{E}\big[\| \bXq^{\top}\big(\bar{s}(\bXq,\bWq^{(t)})-\hat{s}(\bXq\times\bw^{(t)})\big) \|^{2}_{2} \big]  \notag\\
&= \frac{1}{m^2}\mathbb{E}\big[ Tr\big( \bXq^{\top} \bq^{(t)}\bq^{(t)\top} \bXq \big) \big]  \notag\\
&= \frac{1}{m^2} Tr\big( \bXq^{\top} \mathbb{E}[\bq^{(t)}\bq^{(t)\top}] \bXq \big), \label{eq:var1}
\end{align}
where $Tr(\cdot)$ denotes the trace of a matrix, and we let $\bq^{(t)}\triangleq \bar{s}(\bXq,\bWq^{(t)})-\hat{s}(\bXq \times\bw^{(t)})$.
From Lemma 4 of~\cite{ZipML_arXiv}, we have that
\begin{equation}\label{eq:var2}
    \mathbb{E}\big[q_i^{(t)}q_j^{(t)}]
    \left\{
    \begin{array}{ll}
          \leq 2^{-2l_w}\Big(\sum_{k=0}^{r}c_k\big(\bxq_i\bw^{(t)}\big)^{k}\Big)^2& \text{if \; } i\!=\!j\\
          =0 & \text{\!\!otherwise},
    \end{array} 
    \right.
\end{equation}
where $q_i^{(t)}$ denotes the $i^{th}$ element of $\bq^{(t)}$.
Combining equations \eqref{eq:var1} and \eqref{eq:var2} with the fact that  $\Big(\sum_{k=0}^{r}c_k\big(\bxq_i\bw^{(t)}\big)^{k}\Big)^2 \approx \big( {s}(\bxq_i\cdot \bw^{(t)})\big)^2 \leq 1$ for all $i\in[m]$, we obtain 
% and combining this assumption with ~\eqref{eq:var1} and ~\eqref{eq:var2}, we find that
\begin{equation*}
    \mathbb{E}\big[\| \mathbf{p}^{(t)} - \mathbb{E}[\mathbf{p}^{(t)}] \|^{2}_{2} \big] 
    \leq \frac{1}{2^{-2l_w}m^2} Tr\big( \bXq^{\top} \bXq \big) =\frac{1}{2^{-2l_w}m^2}\| \bXq \|_{F}^{2}.
\end{equation*}

\subsection{Proof of Lemma~\ref{lemma:lipschitz}}\label{app:lipschitz_proof}
For the logistic regression cost function $C(\mathbf{w})$, the Lipschitz constant $L$ is less than or equal to the largest eigenvalue of the Hessian $\nabla^{2}(\bw)$ for all $\bw$ and is given by
\begin{equation}
    L = \frac{1}{4}\max\big\{\text{eig}\big(\bXq^{\top}\bXq\big)\big\}.
\end{equation}

\subsection{Proof of Theorem~\ref{thm:privacy}}\label{app:mainthm_proof}
(Convergence) 
First, we show that the master can decode $\bXq^{\top}\bar{s}(\bXq,\bWq^{(t)})$ over the finite field as long as $N\geq (2r+1)(K+T-1) +1$.
As described in Section~\RNum{3}, given the polynomial approximation of the sigmoid function  in (13), the degree of $h(z)$ in (19) is at most $(2r+1)(K+T-1)$. The decoding process uses the computations from the workers as evaluation points $h(\alpha_i)$ to interpolate the polynomial $h(z)$. 
% As described in Section~\RNum{3}, given the polynomial approximation of the sigmoid function  in~\eqref{eq:poly_approximation}, the degree of $h(z)$ in~\eqref{eq:beta} is at most $(2r+1)(K+T-1)$. The decoding process uses the computations from the workers as evaluation points $h(\alpha_i)$ to interpolate the polynomial $h(z)$. 
The master can obtain all of the  coefficients of $h(z)$ as long as it collects at least $\text{deg}\big( h(z) \big) + 1 \leq (2r+1)(K+T-1)+1$ evaluation results of $h(\alpha_i)$. 
After $h(z)$ is recovered, the master can decode the sub-gradient $\overline{\mathbf{X}}_i^{\top}\bar{s}(\overline{\mathbf{X}}_i, \bWq^{(t)})$ by computing $h(\beta_i)$ for  $i\in[K]$. Hence, the recovery threshold is given by $(2r+1)(K+T-1) +1$ to decode $\bXq^{\top}\bar{s}(\bXq,\bWq^{(t)})$.

% Next, we consider the update equation in {\cpml} (see~\eqref{eq:updateEQN}) and prove its convergence to $\bw^{*}$. 
% From the $L$-Lipschitz continuity of $\nabla C({\bw})$ stated in Lemma~\ref{lemma:lipschitz}, we have 
Next, we consider the update equation in {\cpml} (see (25)) and prove its convergence to $\bw^{*}$. 
From the $L$-Lipschitz continuity of $\nabla C({\bw})$ stated in Lemma 2, we have  
\begin{align*}
    C(\bw^{(t+1)}) &\!\leq \!C(\bw^{(t)}) \!+\! \langle \nabla C(\bw^{(t)}), \bw^{(t+1)}\!-\!\bw^{(t)}\rangle 
    \!+\! \frac{L}{2}{\| \bw^{(t+1)}\!-\!\bw^{(t)}\! \|}^2  \\
    &\!=\! C( \bw^{(t)} )-\eta \langle \gC ( \bw^{(t)} ), \mathbf{p}^{(t)} \rangle + \frac{L}{2}{\| \mathbf{p}^{(t)} \|}^2,
\end{align*} 
where $\langle ,\cdot, \rangle$ is the inner product \cite{boyd2004convex}.% \cite{boyd2004convex}.
By taking the expectation with respect to the quantization noise on both sides, 
\begin{align}
    &\mathbb{E}\big[ C(\bw^{(t+1)}) \big] \notag \\
    &\leq C( \bw^{(t)} ) - \eta \| \gC ( \bw^{(t)} )\|^2 +\frac{L\eta^2}{2} \big( \| \gC ( \bw^{(t)} )\|^2 + \sigma^2 \big) \notag \\
    &\leq C( \bw^{(t)} ) - \eta \big(1-\frac{L\eta}{2}\big)\| \gC ( \bw^{(t)} )\|^2+\frac{L\eta^2\sigma^2}{2} \notag \\
    &\leq C( \bw^{(t)} ) - \frac{\eta}{2}\| \gC ( \bw^{(t)} )\|^2\!+\!\frac{\eta\sigma^2}{2} \label{eqn1}\\
    &\leq C(\bw^{*})\!+\!\langle \gC ( \bw^{(t)} ), \bw^{(t)} \!-\! \bw^{*}\rangle 
    - \frac{\eta}{2} \| \gC ( \bw^{(t)} )\|^2\!+\!\frac{\eta\sigma^2}{2} \label{eqn2}\\
    &\leq C(\bw^{*})+ \langle\mathbb{E}[\mathbf{p}^{(t)}],\bw^{(t)} - \bw^{*}\rangle - \frac{\eta}{2} \mathbb{E}\big[\| \mathbf{p}^{(t)} )\|^2\big]+\eta\sigma^2 \label{eqn3}\\
    &= C(\bw^{*}) +\eta\sigma^2 +\mathbb{E}\Big[ \langle \mathbf{p}^{(t)},\bw^{(t)}-\bw^{*} \rangle-\frac{\eta}{2}\| \mathbf{p}^{(t)} )\|^2\Big] \notag \\
    &= C(\bw^{*}) +\eta\sigma^2 + \frac{1}{2\eta}\big( \mathbb{E}\| \bw^{(t)} - \bw^{*} \|^2 -\mathbb{E}\| \bw^{(t+1)} - \bw^{*} )\|^2\big), \notag 
    % &= C(\bw^{*}) +\eta\sigma^2 + \frac{1}{2\eta}\big( \| \bw^{(t)} - \bw^{*} \|^2 -\| \bw^{(t+1)} - \bw^{*} )\|^2\big) \notag 
\end{align}
where \eqref{eqn1} follows from $L\eta\leq1$, 
\eqref{eqn2} from the convexity of $C$, and
\eqref{eqn3} holds since $\mathbb{E}[\mathbf{p}^{(t)}]=\gC ( \bw^{(t)} )$ and $\mathbb{E} \big[\| \mathbf{p}^{(t)} )\|^2 \big] - \| \gC ( \bw^{(t)} )\|^2 \leq \sigma^2$ from Lemma 1 by assuming an arbitrarily large $r$. Summing the above equations for $t=0,\ldots,J-1$, we have
 \begin{align*}
    \sum_{t=0}^{J-1}\Big(\mathbb{E}\big[ C(\bw^{(t+1)})\big]-C(\bw^{*})\Big) %\notag \\
    & \leq \frac{1}{2\eta}\big( \mathbb{E}\| \bw^{(0)} - \bw^{*} \|^2 - \mathbb{E}\| \bw^{(J)} - \bw^{*} )\|^2\big) + J\eta\sigma^2\\
    & \leq \frac{\| \bw^{(0)} - \bw^{*} \|^2}{2\eta} + J\eta\sigma^2.
\end{align*}
Finally, since $C$ is convex, we observe that, 
\begin{align*}
    \mathbb{E}\Big[ C\big( \frac{1}{J}\sum_{t=0}^{J}\bw^{(t)}\big)\Big]\!-\!C(\bw^{*})
    &\leq \frac{1}{J}\sum_{t=0}^{J-1}\Big(\mathbb{E}\big[ C(\bw^{(t+1)})\big]\!-\!C(\bw^{*})\Big) \\
    &\leq \frac{\| \bw^{(0)} - \bw^{*} \|^2}{2\eta J} + \eta\sigma^2,%, 
\end{align*}
which completes the proof of convergence.

(Privacy) 
% Proof of privacy is deferred to Appendix~\ref{app:pravicyproof}.
% Let $\mathbf{U}^{top}\in\fF_p^{K\times N}$ and $\mathbf{U}^{bottom}\in\fF_p^{T\times N}$ are the top and bottom submatrix of the encoding matrix $\mathbf{U}$ constructed in Section~\ref{section:proposed_scheme}, respectively. 
Let $\mathbf{U}^{top}\in\fF_p^{K\times N}$ and $\mathbf{U}^{bottom}\in\fF_p^{T\times N}$ are the top and bottom submatrix of the encoding matrix $\mathbf{U}$ constructed in Section~\RNum{3}, respectively. 
From Lemma 2 of \cite{yu2018lagrange}, $\mathbf{U}^{bottom}$ is an MDS matrix. Therefore, every $T\times T$ submatrix of $\mathbf{U}^{bottom}$ is invertible.

For a colluding set of workers $\mathcal{T}\subset{[N]}$ of size $T$, their received dataset satisfies, 
\begin{equation}\label{eq:X_tilde_T}
\widetilde{\bX}_{\mathcal{T}}=\overline{\bX}\times \mathbf{U}^{top}_{\mathcal{T}}+\mathbf{R}\times \mathbf{U}^{bottom}_{\mathcal{T}},
\end{equation}
where $\mathbf{R}=\big(\mathbf{R}_{K},\ldots,\mathbf{R}_{K+T}\big)$, and 
$\mathbf{U}^{top}_{\mathcal{T}}\in\fF_p^{K\times T}$ and $\mathbf{U}^{bottom}_{\mathcal{T}}\in\fF_p^{T\times T}$ are the top and bottom submatrices which correspond to the columns in $\mathbf{U}$ that are indexed by $\mathcal{T}$. All elements of $\mathbf{R}$  are independent and uniformly distributed over the finite field $\fF_{p}$.

Similarly, $\widetilde{\bW}^{(t)}_\mathcal{T}$ can be represented as 
\begin{equation}\label{eq:W_tilde_T}
    \widetilde{\bW}^{(t)}_\mathcal{T}=\overline{\bW}^{(t)}\times \mathbf{U}^{top}_{\mathcal{T}}+\mathbf{V}^{(t)}\times \mathbf{U}^{bottom}_{\mathcal{T}},
\end{equation}
where $\mathbf{V}^{(t)}=(\mathbf{V}^{(t)}_{K+1},\ldots,\mathbf{V}^{(t)}_{K+T})$ and all elements of $\mathbf{V}^{(t)}$ are independent and uniformly distributed over the finite field $\fF_{p}$, for all $t\in[J]$. 
We first show that $I(\bXtildeT; \overline{\bX})$ =0 as follows. 
\begin{align}
    I(\bXtildeT; \overline{\bX})
    &= H( \bXtildeT ) - H( \bXtildeT | \overline{\bX}) \notag\\
    &= H( \bXtildeT ) - H( \overline{\bX} \mathbf{U}^{top}_{\mathcal{T}}+\mathbf{R} \mathbf{U}^{bottom}_{\mathcal{T}} | \overline{\bX}) \label{MI_proof_1}\\
    &= H( \bXtildeT ) - H( \mathbf{R} \mathbf{U}^{bottom}_{\mathcal{T}} | \overline{\bX}) \label{MI_proof_2} \\
    &= H( \bXtildeT ) - H( \mathbf{R} ) \label{MI_proof_3}\\
    &= 0, \label{MI_proof_4}
\end{align}
where~\eqref{MI_proof_1} follows from~\eqref{eq:X_tilde_T},~\eqref{MI_proof_2} holds since we can drop $\overline{\bX} \mathbf{U}^{top}_{\mathcal{T}}$ given $\overline{\bX}$ as the former is a deterministic function of the latter. Equation~\eqref{MI_proof_3} holds since $\mathbf{R}$ and $\overline{\bX}$ are independent and $\mathbf{U}^{bottom}_{\mathcal{T}}$ is invertible. Equation~\eqref{MI_proof_4} holds from the observation that $\mathbf{R}$ is a uniformly distributed random matrix, hence it has the  maximum entropy in the finite field  $\fF_p$, combined with the fact that mutual information is always non-negative. 
Next, we prove $I\big(\overline{\bX} ; \widetilde{\mathbf{X}}_\mathcal{T}, \{\bWtildeT\}_{t\in[J]}\big) = 0$. We first obtain
\begin{align}
    &I\Big(\overline{\bX} ; \bXtildeT, \{\bWtildeT\}_{t\in[J]}\Big) \notag\\
    &\quad = I\big(\overline{\bX};\bXtildeT\big) + I\Big( \overline{\bX} ; \{\bWtildeT\}_{t\in[J]} \big| \bXtildeT\Big)\notag\\
    &\quad = I\Big( \overline{\bX} ; \{\bWtildeT\}_{t\in[J]} \big| \bXtildeT\Big) \label{MI_proof_5}\\
    &\quad = H\Big( \{\bWtildeT\}_{t\in[J]} \big| \bXtildeT\Big) - H\Big( \{\bWtildeT\}_{t\in[J]} \big| \overline{\bX},\bXtildeT\Big) \notag \\
    &\quad = \sum_{j=1}^{J} \Bigg ( 
    H\Big( \widetilde{\mathbf{W}}^{(j)}_\mathcal{T} \big| \{\bWtildeT\}_{t\in[j-1]}, \bXtildeT\Big) - H\Big( \widetilde{\mathbf{W}}^{(j)}_\mathcal{T} \big| \{\bWtildeT\}_{t\in[j-1]},\overline{\bX},\bXtildeT\Big)\Bigg), \label{MI_proof_6}
\end{align}
where~\eqref{MI_proof_5} follows from \eqref{MI_proof_4}, and~\eqref{MI_proof_6} from the chain rule of entropy. 
% Since $\widetilde{\bX}_\mathcal{T}$ and $\widetilde{\bW}^{(t)}_\mathcal{T}$ are completely masked by the random padding matrices, T colluding workers get no information about $\overline{\bX}$, i.e., $I\Big(\overline{\bX} ; \widetilde{\mathbf{X}}_\mathcal{T}, \{\widetilde{\mathbf{W}}^{(t)}_\mathcal{T}\}_{t\in[J]}\Big) = 0$. 
From the second term of~\eqref{MI_proof_6}, we derive
\begin{align}
    % &H\Big( \widetilde{\mathbf{W}}^{(j)}_\mathcal{T} \big| \{\bWtildeT\}_{t\in[j-1]},\overline{\bX},\bXtildeT\Big)  \notag \\
    % &\qquad\geq H\Big( \widetilde{\mathbf{W}}^{(j)}_\mathcal{T} \big| \overline{\bW}^{(j)}, \{\bWtildeT\}_{t\in[j-1]},\overline{\bX},\bXtildeT\Big) \label{MI_proof_7} \\
    % &\qquad= H\Big( \widetilde{\mathbf{W}}^{(j)}_\mathcal{T} \big| \overline{\bW}^{(j)} \Big) \label{MI_proof_8} \\
    % &\qquad= H\Big( \overline{\bW}^{(j)} \mathbf{U}^{top}_{\mathcal{T}}+\mathbf{V}^{(j)} \mathbf{U}^{bottom}_{\mathcal{T}} \big| \overline{\bW}^{(j)} \Big) \label{MI_proof_9}\\
    % &\qquad= H\big( \mathbf{V}^{(j)}\big) \label{MI_proof_10}
    H\Big( \widetilde{\mathbf{W}}^{(j)}_\mathcal{T} \big| \{\bWtildeT\}_{t\in[j-1]},\overline{\bX},\bXtildeT\Big)  
    &\geq H\Big( \widetilde{\mathbf{W}}^{(j)}_\mathcal{T} \big| \overline{\bW}^{(j)}, \{\bWtildeT\}_{t\in[j-1]},\overline{\bX},\bXtildeT\Big) \label{MI_proof_7} \\
    &= H\Big( \widetilde{\mathbf{W}}^{(j)}_\mathcal{T} \big| \overline{\bW}^{(j)} \Big) \label{MI_proof_8} \\
    &= H\Big( \overline{\bW}^{(j)} \mathbf{U}^{top}_{\mathcal{T}}+\mathbf{V}^{(j)} \mathbf{U}^{bottom}_{\mathcal{T}} \big| \overline{\bW}^{(j)} \Big) \label{MI_proof_9}\\
    &= H\big( \mathbf{V}^{(j)}\big), \label{MI_proof_10}
\end{align}
where~\eqref{MI_proof_7} holds since  conditioning cannot increase  entropy. Equation~\eqref{MI_proof_8} holds since $\widetilde{\mathbf{W}}^{(j)}_\mathcal{T}$ and $\big (\{\bWtildeT\}_{t\in[j-1]},\overline{\bX},\bXtildeT\big )$ are conditionally independent given $\overline{\bW}^{(j)}$. Equation~\eqref{MI_proof_9} follows from~\eqref{eq:W_tilde_T}, and~\eqref{MI_proof_10} follows from the same steps~\eqref{MI_proof_1}-\eqref{MI_proof_3}.
From~\eqref{MI_proof_6} and ~\eqref{MI_proof_10} we obtain
\begin{align}
    % &I\Big(\overline{\bX} ; \bXtildeT, \{\bWtildeT\}_{t\in[J]}\Big)  \notag \\
    % &\qquad \leq \sum_{j=1}^{J} \left (
    % H\Big( \widetilde{\mathbf{W}}^{(j)}_\mathcal{T} \big| \{\bWtildeT\}_{t\in[j-1]}, \bXtildeT\Big) - H\big( \mathbf{V}^{(j)} \big)\right )= 0 \label{MI_proof_11}
    I\Big(\overline{\bX} ; \bXtildeT, \{\bWtildeT\}_{t\in[J]}\Big)  
    \leq \sum_{j=1}^{J} \left (
    H\Big( \widetilde{\mathbf{W}}^{(j)}_\mathcal{T} \big| \{\bWtildeT\}_{t\in[j-1]}, \bXtildeT\Big) - H\big( \mathbf{V}^{(j)} \big)\right )= 0, \label{MI_proof_11}
\end{align}
where~\eqref{MI_proof_11} holds since $\mathbf{V}^{(j)}$ is a uniformly distributed random matrix, and therefore has the maximum entropy in the finite field $\fF_p$, $H\big( \mathbf{V}^{(j)} \big) \geq H\Big( \widetilde{\mathbf{W}}^{(j)}_\mathcal{T} \big| \{\bWtildeT\}_{t\in[j-1]}, \bXtildeT\Big)$, 
% \begin{equation}
% H\big( \mathbf{V}^{(j)} \big) \geq H\Big( \widetilde{\mathbf{W}}^{(j)}_\mathcal{T} \big| \{\bWtildeT\}_{t\in[j-1]}, \bXtildeT\Big),
% \end{equation}
combined with the fact that mutual information is always non-negative, $I\big(\overline{\bX} ; \bXtildeT, \{\bWtildeT\}_{t\in[J]}\big)\geq 0$. 
Finally, from the data-processing inequality \cite{cover2012elements},
\begin{equation}
I\big(\overline{\bX} ; \widetilde{\mathbf{X}}_\mathcal{T}, \{\widetilde{\mathbf{W}}^{(t)}_\mathcal{T}\}_{t\in[J]}\big) \geq I\big(\bX ; \widetilde{\mathbf{X}}_\mathcal{T}, \{\widetilde{\mathbf{W}}^{(t)}_\mathcal{T}\}_{t\in[J]}\big) \geq 0. 
% 0&\leq I\Big(\bX ; \widetilde{\mathbf{X}}_\mathcal{T}, \{\widetilde{\mathbf{W}}^{(t)}_\mathcal{T}\}_{t\in[J]}\Big)  \\
% &\leq I\Big(\overline{\bX} ; \widetilde{\mathbf{X}}_\mathcal{T}, \{\widetilde{\mathbf{W}}^{(t)}_\mathcal{T}\}_{t\in[J]}\Big) = 0
\end{equation}
Therefore,  $I\big({\bX} ; \widetilde{\mathbf{X}}_\mathcal{T}, \{\widetilde{\mathbf{W}}^{(t)}_\mathcal{T}\}_{t\in[J]}\big) = 0$ and the original dataset remains information-theoretically private against $T$ colluding workers.

\subsection{Details of the Secure Multi-Party Computation (MPC) Implementation}\label{app:BGW}
Our benchmarks are based on two well-known MPC protocols, the notable BGW protocol from \cite{ben1988completeness}, and the more recent MPC protocol from \cite{beerliova2008perfectly, damgaard2007scalable}. 
Both protocols allow computations of polynomial functions, which consists of addition and multiplication operations, in a privacy preserving manner by untrusted workers. 
% To avoid revealing any information about the original values, computations for each operation are carried out over the secret shares. 
At the end of the computation, any collusion between $T$ out of $N$ workers does not reveal any information (in an information-theoretic sense) about the input variables while workers only learn a secret share of the actual result. 
The former protocol is more communication-intensive than the latter, as it incurs a communication cost that is quadratic in the number of workers.
The latter protocol enables the communication cost to scale linearly with respect to the number of workers, however, as a trade-off, it requires a significant amount of offline computations as well as storage load at each worker.

For constructing the secret shares, we use Shamir's secret sharing protocol \cite{shamir1979share}, which protects the privacy of secret variables against any collisions between up to $T$ workers. 
This is done by embedding a given secret $a$ in a degree $T$ polynomial $h(\xi) = a + \xi r_1, \ldots, \xi^T r_T $ where $r_i$, $i\in [T]$ are uniformly random variables. 
The secret share of $a$ at worker $i\in [N]$ is represented by $h(i) = [a]_i$. 
% In the following, we consider two variables $a$ and $b$ that are secret shared among $N$ workers. 
Then, addition and multiplication operations are computed as follows.   

% \noindent
{\bf Addition. } To compute a secure addition $a+b$, workers locally add their secret shares $[a]_i+[b]_i$ and perform a modulo operation. 

% \noindent
{\bf Multiplication. } To compute a secure multiplication $ab$, 
the two protocols differ in their approaches. 
In the BGW protocol from \cite{ben1988completeness}, each worker first  multiplies its secret shares $[a]_i$ and $[b]_i$ locally. 
The resulting value $[a]_i[b]_i$ is a secret share of $ab$, however, the corresponding polynomial has degree $2T$, twice the degree of the original polynomial. This causes the degree  to grow excessively as more multiplication gates are executed.
To alleviate this problem, workers perform a degree-reduction step by creating new shares corresponding to a polynomial of degree $T$, reducing the degree from $2T$.  
The communication overhead of this protocol is $O(N^2)$. 
The protocol from \cite{beerliova2008perfectly, damgaard2007scalable} utilizes offline computations to reduce the communication overhead. In the offline phase, this protocol creates a random variable $\rho$ and two secret shares corresponding to random polynomials with degree $T$ and $2T$, which are denoted by $[\rho]_{T, i}$ and $[\rho]_{2T, i}$, respectively, for worker $i\in[N]$. 
In the online phase, worker $i\in[N]$ locally the multiplies $[a]_i$ with $[b]_i$. Each worker now holds a secret share of $ab$, however, the corresponding polynomial for the secret shares has degree $2T$. 
Worker $i\in[N]$ then locally computes $[a]_i [b]_i - [\rho]_{2T, i}$.   
Next, workers broadcast their local computations to others, after which each worker decodes $ab-\rho$.
Note that the privacy of  $ab$ is still protected since it is masked by the random value $\rho$.  
Finally, each worker locally computes $ab-\rho+ [\rho]_{T, i}$. As a result, $\rho$ cancels out and workers obtain a secret share of $ab$ embedded in a degree $T$ polynomial. 
The communication overhead of this protocol is $O(N)$. For the details, we refer to \cite{beerliova2008efficient}.

\end{document}